\documentclass{article}

\usepackage{arxiv}

\usepackage[utf8]{inputenc} 
\usepackage[T1]{fontenc}  
\usepackage{hyperref}   
\usepackage{url}           
\usepackage{booktabs}      
\usepackage{amsfonts}       
\usepackage{nicefrac}      
\usepackage{microtype}      
\usepackage{graphicx}
\usepackage{natbib}
\usepackage{doi}
\usepackage{multirow}
\usepackage{amsmath}
\usepackage{MnSymbol}
\usepackage{graphicx}
\usepackage{epstopdf}
\usepackage{arydshln}

\title{A survey of neural-network-based methods utilising comparable data for finding translation equivalents}

\author{ \href{https://orcid.org/0009-0001-8402-504X}{\includegraphics[scale=0.06]{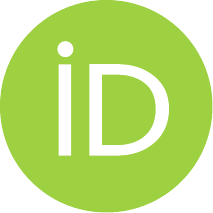}\hspace{1mm}Michaela Denisová} \\
	Natural Language Processing Centre\\
	Faculty of Informatics\\
    Masaryk University \\
	Brno, Czech Republic \\
	\texttt{449884@mail.muni.cz} \\
	\And
	\href{https://orcid.org/0000-0001-5097-4610}{\includegraphics[scale=0.06]{orcid.pdf}\hspace{1mm}Pavel Rychlý} \\
	Lexical Computing \& \\ 
    Natural Language Processing Centre\\
	Faculty of Informatics\\
    Masaryk University \\
	Brno, Czech Republic \\
	\texttt{pary@fi.muni.cz} \\
}

\date{}

\renewcommand{\headeright}{}
\renewcommand{\undertitle}{}
\renewcommand{\shorttitle}{}

\begin{document}
\maketitle

\begin{abstract}
	The importance of inducing bilingual dictionary components in many natural language processing (NLP) applications is indisputable. However, the dictionary compilation process requires extensive work and combines two disciplines, NLP and lexicography, while the former often omits the latter. In this paper, we present the most common approaches from NLP that endeavour to automatically induce one of the essential dictionary components, translation equivalents and focus on the neural-network-based methods using comparable data. We analyse them from a lexicographic perspective since their viewpoints are crucial for improving the described methods. Moreover, we identify the methods that integrate these viewpoints and can be further exploited in various applications that require them. This survey encourages a connection between the NLP and lexicography fields as the NLP field can benefit from lexicographic insights, and it serves as a helping and inspiring material for further research in the context of neural-network-based methods utilising comparable data.
\end{abstract}


\section{Introduction}

Bilingual dictionaries are one of the most crucial learner’s materials when acquiring a foreign language. For a bilingual dictionary, it is important to be comprehensive and include all relevant information useful for its users. However, compiling such an extensive bilingual dictionary requires the work of many lexicographers, and it is a time-consuming and broad process. Involving automatic methods offers a solution to this problem. This paper outlines the recent fully automatic methods and algorithms for extracting one of the most important bilingual dictionary components, translation equivalents.

Finding translation equivalents is also an important task in the NLP field that can be utilised for many applications, such as unsupervised machine translation \citep{Conneau2017WordTW, Artetxe2018UnsupervisedNM, artetxe-etal-2019-effective, duan-etal-2020-bilingual}, cross-lingual information retrieval \citep{lavrenko2002}, cross-lingual dependency parsing \citep{Xiao2014DistributedWR, guo2015parsing}, document classification \citep{klem2012}, language acquisition and learning \citep{yuan-etal-2020-interactive}, etc. However, researchers in the NLP field are not often aware of lexicography-related issues that are crucial for improving their models. Instead, they focus on plain word-to-word extraction without considering any natural language engineering aspects, gaining incorrect results. For example, the research conducted by \cite{denisova-rychly-2023} shows that evaluation produces different outcomes when natural language engineering aspects are reflected in the evaluation datasets and processes (see Table~\ref{tab:article}).

\begin{table}
\caption{Partial table from the \cite{denisova-rychly-2023} displaying comparison between the results when using the MUSE dataset (broadly used among NLP researchers) and dictionary data compiled by lexicographers for evaluating two supervised benchmark cross-lingual embedding models trained on two monolingual word embeddings (MEs) from different type of data for English-Korean.}
\centering
    \begin{tabular}{ccccc}
    \toprule
    \multicolumn{1}{c}{\multirow{2}{1cm}{(\%)}} & \multicolumn{2}{c}{MUSE dataset} & \multicolumn{2}{c}{Dictionary data} \\
    \cmidrule(r){2-5}
    & Wiki MEs & WebCorpus MEs & Wiki MEs & WebCorpus MEs \\
    \midrule
    VecMap & 49.00 & 35.44 & \textbf{29.41} & 36.24 \\
    \midrule
    RCLS & \textbf{51.91} & \textbf{37.60} & 28.37 &	\textbf{37.80} \\
    \bottomrule
    \end{tabular}
    \label{tab:article}
  \end{table}

In Table~\ref{tab:article}, we can observe that evaluating the same models with dictionary data instead of automatically compiled data gives us drastically different results, dropping by a margin of almost 20\%, and it changes the models' ranking (i.e., which model is better). This implies two conclusions. Firstly, the actual models' performance is worse than the papers state when evaluated on correct, manually compiled data by language experts. Secondly, many models have been evaluated and optimised for erroneous data, yielding inaccurate results and preventing genuine advancements. In other words, the progress and comparisons between the models that the papers state cannot be trusted. Therefore, in this survey, we show the most vital lexicographic viewpoints that are beneficial for research in NLP since considering and implementing them into the models can bring significant improvements for these applications. We aim this survey to help improve these NLP applications through lexicography as we observed that many problems result from neglecting the lexicographic side of the problem.

In lexicography, computers have been involved in the dictionary compiling process since the sixties of the last century. Initially, dictionaries were converted into machine-readable lexical databases \citep{fontenelle1997}. Later, the corpus data was employed as a basis for lexicographic information. Over the years, the role of computers and electronic methods has increased. Nowadays, lexicographers rely on software and tools provided by the NLP community \citep{atkins2008oxford}. Modern approaches in the NLP field develop quickly, and recent trends suggest the usage of fully automatic methods involving neural networks.

Such recent trends are also incorporated in modern lexicography and dictionaries compilation \citep{jakubivcek2021million}. However, research in the NLP that could be utilised in lexicography often concentrates on the technical side of the problem and does not involve any discussion from a lexicographic point of view. Moreover, researchers in the NLP field are not often aware of lexicography-related issues that are crucial for improving their models.

In this paper, we group the automatic methods for finding translation equivalents into parallel- and comparable-data-based. Further, both categories contain the following subcategories: statistical-based and neural-network-based methods. We focus mainly on neural-network-based methods using comparable data. The reasons behind this are the following. Firstly, these methods have enjoyed great popularity among researchers who re-focused their attention from statistical-based to neural-network-based methods \citep{zhang-etal-2017-adversarial}, given the immense number of recent publications \cite[etc.]{jawanpuria-etal-2020-geometry, Cao2021WordET, wang-etal-2021-multi, severini-etal-2022-dont} and competitive results \citep{Mikolov2013ExploitingSA, zhang-etal-2017-adversarial, Conneau2017WordTW}. Secondly, having comparable data is a more realistic scenario for various language pairs and combinations \citep{Fung1998ASV, zhang-etal-2017-adversarial}, and they are more diverse in their texts \citep{atkins2008oxford}.    

We comprehensively survey the neural network-based methods using comparable data and analyse them from a lexicographic point of view. We point out problematic natural language engineering aspects often omitted in the current NLP research of finding translation equivalents. These aspects are key for improving NLP applications that incorporate translation equivalents.

Our motivation for this paper is to thoroughly map, classify, and describe currently used comparable-data-based algorithms that utilise neural networks. We aim to capture recent advances from the NLP field and situate them in the context of lexicography. This survey links the needs of NLP with the insights that lexicography offers. It brings a valuable lexicographic perspective into the NLP, which is essential for improving these methods and many useful applications. 

Our contribution is the following:
\begin{enumerate}
    \item We define the most important natural language engineering aspects to consider when extracting translation equivalents via neural-network-based methods using comparable data and explore how the algorithms handle these aspects.
    \item We outline the most common data types and approaches for extracting translation equivalents and state their advantages and disadvantages.
    \item We provide a comprehensive overview of neural-network-based methods using comparable data, track and outline the important aspects we defined in (1), and reason their importance and usage in this field.
    \item We point out problems concerning the lexicographic side of finding translation equivalents automatically, which are valuable for NLP to improve the models' performance. 
\end{enumerate}

This paper is structured as follows. Section~\ref{sec:treeq} outlines the natural language engineering aspects of finding translation equivalents to incorporate them into the methods’ analysis. Sections~\ref{sec:par} and~\ref{sec:comp} discuss parallel- and comparable-data-based methods and state their benefits and drawbacks. Sections~\ref{sec:static} and~\ref{sec:dynamic} classify and describe the comparable-data-based methods, focusing on the aspects indicated in Section~\ref{sec:treeq}. Section~\ref{sec:discuss} presents the methods from Sections~\ref{sec:static} and~\ref{sec:dynamic} in the detailed summary tables~\ref{tab:1}-~\ref{tab:4} and critically examines how well the methods can help with the translation equivalents retrieval task. Section~\ref{sec:conc} offers concluding remarks and proposes suggestions for future works. 

\section{Translation equivalents}
\label{sec:treeq}

In this Section, we introduce the crucial natural language engineering aspects to define and consider when dealing with the translation equivalents retrieval task. This step is necessary to identify the aspects to focus on.

One of the most significant parts of the bilingual dictionary is the translation equivalents of the dictionary entries. However, the translation equivalent is not always straightforward to determine, and it can cause doubts even among lexicographers or translators. Moreover, not every headword from the source language has a direct translation into the target language. For example, the German word \emph{abhärten} does not have a direct translation into English, but we can use the activity description instead: \emph{cold water immersion}, \emph{to get used to the cold}, \emph{cold water therapy}, \emph{cold plunge}, etc. 

Thus, it is important to include not only direct translation equivalents but also near-equivalents, contextual translations, and glosses to avoid this problem and broaden the definition of translation equivalent. 

Furthermore, the problem is that sometimes we cannot translate one word with only one word. For example, the German word \emph{Grundschule} translates to English as \emph{elementary school} or \emph{primary school}. In this case, the translation equivalent consists of multiple words called multi-word expressions. Among others, multi-word expressions include phrases, idioms or multiple words that carry one sense. Handling multi-word expressions is important when retrieving translation equivalents; therefore, we factor this into the survey. 

Another questionable aspect of retrieving translation equivalents automatically is that it is not always clear what word form we want the model to generate, especially in morphologically rich languages. For example, the English word \emph{fluid} has in Estonian many forms, i.e., \emph{vedelik}, \emph{vedeliku}, \emph{vedelikku}, etc. The issue lies in whether we want the model to find the basic word form (\emph{vedelik}), even though it is not the most common one, and there is a lack of evidence of this form in the available data. Or is it more reasonable to extract all/ the most common inflected word forms? And if we decide on one or another, should we penalise the model for finding the opposite?

When generating translation equivalents, we should consider polysemic words with multiple senses. For instance, the English noun \emph{bank} can be translated to German as \emph{die Bank} (financial institution), as \emph{der Wall} (a long raised mass of earth), as \emph{die Ufer} (side of the river), etc. Do we want the model to find translation equivalents in the target language of all possible source words’ senses? And if we decide which translation equivalents we want, should we penalise the model for finding the other ones?

Equally important is to reflect the language pairs for which the method is suitable. Some methods may work for big language pairs with a great amount of data available but perform poorly for low-resource languages. The same applies to other natural language engineering aspects. For example, when considering only analytical languages such as English or Chinese, there is no need for a method that accounts for morphological variance. However, with these exceptions aside, when our scope broadens, considering morphological variance becomes almost invariably necessary. Additionally, the linguistic distance between the source and target language influences the quality of the extracted translation equivalents \citep{sogaard-etal-2018-limitations}. 

The last essential aspect of retrieving translation equivalents is their final purpose and users. This information guides the selection of optimal methods tailored to specific objectives. Some articles indicate their importance in the unsupervised machine translation approaches as a way to handle out-of-vocabulary words or in other NLP applications, such as cross-lingual named entity recognition, cross-lingual information retrieval, language acquisition, etc. Nonetheless, in many works, finding translation equivalents is seen as a standalone task, and their further usage is not usually discussed. Moreover, application in the lexicography field is barely considered. 

These aspects are often omitted in the articles that deal with translation equivalents' extraction. The methods focus mainly on different alignment strategies and the computational side of the problem. The NLP researchers often lack a deeper understanding of how language works, treating word-to-word translations in isolation and not as a part of a broader language system. However, we consider them essential when improving the methods that retrieve translation equivalents. Thus, we want to examine if and how the automatic methods handle these questionable points. If the method stands out in one of these aspects, we mention it in the main part of this survey, and we provide a detailed list of all these aspects for each method in the summary tables~\ref{tab:1}-~\ref{tab:4}.

We classify the methods for retrieving translation equivalents into parallel-data- and comparable-data-based, using either statistical or neural network approaches. This survey focuses mainly on neural-network-based methods with comparable data. However, we find it important to list the alternatives and put them in contrast with our chosen method. Therefore, the following sections briefly discuss alternate approaches.

\section{Parallel-data-based methods}
\label{sec:par}

The most common way to automatically generate translation equivalents is through parallel data such as the parallel corpus. A parallel corpus, or in other literature referred to as a translation corpus, is a collection of texts in two or more languages that are equivalent to each other \citep{atkins2008oxford}. In the NLP field, the parallel corpus contains two or more monolingual corpora aligned on the level of words, phrases, or sentences \citep{skengforbillex}.

For many years, using parallel corpus when building a bilingual dictionary has been a preferred method among lexicographers. The reasons favouring parallel corpus are: it contains rich context information, offers many translation equivalents’ candidates, and enables us to extract many-to-many equivalents, including multi-word expressions \citep{atkins2008oxford}. However, rare language combinations or low-resource languages often lack enough parallel data or do not possess any, whereas having a comparable corpus is more feasible \citep{Fung1998ASV}. Additionally, the genres of the parallel texts included in the corpus are often unbalanced and restricted to legislative or law fields \citep{baisa-etal-2016-european}.

One example of a parallel-data-based method is the statistical-based method that computes the probability of the alignment being a translation equivalent using the frequencies of the occurrences as a score \citep{skengforbillex}. Such an approach is usually followed by manual post-editing by lexicographic experts, which is often time-consuming and expensive. Another less common approach is compiling a new language pair dictionary from two existing bilingual dictionaries that share a common language \citep{tanaka-umemura-1994-construction, satoshi, saralegi-etal-2012-building, Ordan2017AutogeneratingBD, denisova}. The drawback of this method is that many word pairs could be incorrectly aligned due to polysemy \citep{denisova}.

A comparative study has been conducted to analyse the aforementioned parallel-data-based statistical approaches in contrast with the neural-network-based methods utilising comparable data \citep{denisova-2022}. The findings of this study revealed that comparable-data-based methods surpassed the ones using parallel data, under some settings by a substantial margin of 30\%. The study suggests that comparable data offer a better option for rare language pairs, supporting their claim by evaluating the systems using the Estonian-Slovak language pair.

However, recent advances in using parallel corpus concentrate mainly on neural machine translation systems (NMTS) \citep{Stahlberg2020NeuralMT}. The great advantage is that they work with the context and can handle sentences, phrases, multi-word expressions, etc. Moreover, they are not limited to the parallel data we possess, but they can produce translation equivalents for various words and sentences they have never seen before, depending on how good the model is \citep{klein-etal-2017-opennmt, tensor2tensor, ott2019fairseq, lewis-etal-2020-bart}. 

The most common approach how to utilise the NMTS for retrieving translation equivalents involves generating target word candidates for each source word. This technique has also been employed to compile several gold-standard evaluation datasets for evaluating the translation equivalents' retrieval systems \cite[etc.]{Dinu2015ImprovingZL, Conneau2017WordTW, glavas-etal-2019-properly, vulic-etal-2019-really} (See Section~\ref{sec:bench}). Retrieving translation equivalents in this way may constrain the ability to generate the multiple senses of a single word since the context is missing. Another approach is to exploit the contextualised embeddings to train the dynamic models thoroughly described in Section~\ref{sec:dynamic} in more detail. We now introduce NMTS, which play a critical role in our later discussion of dynamic methods.

Several approaches that neural machine translation utilises include recurrent neural networks (RNN), convolutional neural networks (CNN), and transformer models, i.e., transformers \citep{Wang2021ProgressIM}. However, transformers have been one of the most trending NMTS over the last few years \cite[etc.]{Vaswani2017AttentionIA, devlin-etal-2019-bert, DBLP:journals/corr/abs-1910-10683, tiedemann-thottingal-2020-opus}.

Like other NMTS, transformers involve context information; they handle inflected word forms and provide translation on the level of words and sentences. However, their training is computationally expensive and requires parallel data that is not always available or balanced in text genres, especially for low-resource languages or rare language combinations. Furthermore, the pre-trained models could be more diverse in language combinations, as they are often paired with English and are available only for high-resource languages.

Among the transformer models often used for machine translation are BART \citep{lewis-etal-2020-bart}, mBART \citep{liu-etal-2020-multilingual-denoising}, MarianMT \citep{tiedemann-thottingal-2020-opus}, and T5 \citep{DBLP:journals/corr/abs-1910-10683}. All of them are available as pre-trained models, for example, in the HuggingFace
library.\footnote{\href{https://huggingface.co/}{https://huggingface.co/}}

BART is a denoising autoencoder for sequence-to-sequence tasks, pre-trained on a corpus consisting of news, books, stories, and web texts for the English language. It was evaluated on the Romanian-English language combination. 
On the contrary, mBART is multilingual BART trained on non-aligned multilingual (Common Crawl web corpus)\footnote{\href{https://commoncrawl.org/}{https://commoncrawl.org/}} and parallel corpus for 24 language pairs (mainly in the public domain), including German, Chinese, Finnish, Estonian, Romanian, etc., all paired with English.

MarianMT was trained using the Marian C++ library\footnote{\href{https://marian-nmt.github.io/}{https://marian-nmt.github.io/}} on OPUS parallel corpora\footnote{\href{https://opus.nlpl.eu/}{https://opus.nlpl.eu/}} \citep{tiedemann-2012-parallel}, which has various domains, such as subtitles, public texts, web texts, etc. It supports various languages and language combinations (not only from and to English), including European, non-European, endangered languages, etc. 

T5 is a text-to-text encoder-decoder that was trained for several tasks, such as summarisation, machine translation, question answering, etc. For machine translation, it was pre-trained for English-French, English-German, and English-Romanian language combinations. Parallel corpora consisted of news, web, and public texts. Currently, the HuggingFace library also has the mT5 \citep{xue-etal-2021-mt5} model (multilingual T5) pre-trained for around 100 languages.

\section{Comparable-data-based methods}
\label{sec:comp}

The opposite of parallel data is comparable data. In this case, we refer to the comparable corpus. A comparable corpus is a set of texts from two or more languages that originate from the same domain or the conditions for their collection were the same \citep{atkins2008oxford}. In the NLP, they are defined as non-aligned texts but similar in genre and size \citep{skengforbillex}. 

Their main advantage compared to parallel data is that they are often available even for low-resource or non-standard language combinations. Parallel corpora are also known to distort the real distribution of items in the target language, pushing up the frequencies of most frequent words and cognates while unfairly under-representing the frequencies of other, sometimes more natural equivalents. Moreover, the variety of texts is usually much wider than in parallel corpora. These are the main reasons that make the comparable-data-based methods our main focus in this paper.

Similarly to the parallel-data-based, comparable-data-based methods can be divided into two subcategories, the ones that use either statistical or neural-network-based approaches.

There has been some research conducted on statistical-based methods utilising comparable data, initially provided by \cite{rapp95, Fung1998ASV}, and further developed by \cite{rapp99, Haghighi2008LearningBL, schafyaro02, kk02, Gaussier2004AGV}. However, recent advances reoriented to neural-network-based approaches, and currently, they produce more research papers than statistical approaches \cite[etc.]{artetxe-etal-2018-robust, kementchedjhieva-etal-2020-generalizing, woller-etal-2021-neglect, Bai2022DualityRF, Marchisio2022BilingualLI}. The reason for this attraction is that they provide competitive results \citep{Mikolov2013ExploitingSA, zhang-etal-2017-adversarial, Conneau2017WordTW}. This motivates us to constrain our survey to neural-network-based methods using comparable data.  

Recent trends in retrieving translation equivalents via neural networks with comparable data prefer cross-lingual embedding models. Cross-lingual embedding models are useful for compiling bilingual dictionaries as they facilitate the transfer of lexical knowledge across languages. The main objective is to learn shared cross-lingual embedding space where similar words obtain similar vectors irrespective of their language \citep{Ruder2017ASO}. Vector representations are used to mathematically express the meaning of the word based on the context words accompanying the word \citep{Smith2019ContextualWR}.

Most cross-lingual embedding models are rooted in monolingual embedding models, or they utilise them directly in the training process \citep{Ruder2017ASO}. Many papers exploit the fastText\footnote{\href{https://fasttext.cc/}{https://fasttext.cc/}}  \citep{bojanowski-etal-2017-enriching} monolingual word embedding model \cite[etc.]{ren-etal-2020-graph, Riley2020UnsupervisedBL, mohiuddin-etal-2020-lnmap, vulic-etal-2020-improving, glavas-vulic-2020-non}, or word2vec \citep{Mikolov2013EfficientEO} skip-gram and CBOW models \cite[etc.]{severini-etal-2020-lmu, michel-etal-2020-exploring, laville-etal-2020-taln}, or a combination of both approaches \citep{eder-etal-2021-anchor}.  

These monolingual word embeddings are often trained on Common Crawl or Wikipedia \cite[etc.]{vulic-etal-2020-improving, chakravarthi-etal-2020-bilingual, Riley2020UnsupervisedBL, marchisio-etal-2021-analysis-euclidean, Sannigrahi2022IsomorphicCE}, or WaCKy corpora \citep{Baroni2009TheWW, laville-etal-2020-taln}, or UN corpus \citep{edmiston-etal-2022-domain}, etc. Sometimes smaller corpora appear, such as Palito corpus \citep{dita-etal-2009-building, michel-etal-2020-exploring}, and many others. Importantly, according to the study by \cite{sogaard-etal-2018-limitations}, the matching domain of the corpora is an essential factor that influences the models’ performance. 

Except for monolingual word embeddings trained on monolingual corpora, they could involve a certain level of supervision in training. Supervision signals are usually in the format of a word-to-word seed lexicon. Based on their size and quality, models are divided into supervised (bigger seed lexicon, usually up to 5K word pairs\footnote{\cite{vulic-korhonen-2016-role} researched that using a seed lexicon with more than 5K word pairs does not significantly influence the resulting quality of the translation equivalents. This was confirmed by \cite{izbicki-2022-aligning}.}) and semi-supervised (smaller seed lexicon or relying on identical strings and numerals). Methods that do not require any supervision signals at all are called unsupervised \citep{Ruder2017ASO}.  

In this field, finding translation equivalents is also known as the bilingual lexicon induction task (BLI) or, in some papers, referred to as bilingual dictionary induction (BDI). Many papers often refer to many-to-one, one-to-many or one-to-one word alignments as a bilingual lexicon or dictionary. However, it has a much broader meaning in the lexicography field, including other bilingual dictionary components (definitions, examples, collocations, etc.). In this paper, we use the term BLI task since it is commonly used in publications.

The BLI task is an intrinsic evaluation task whose objective is to find the most suitable target word \emph{$t_i$} for each query \emph{$s_i$}, provided a list of \emph{N} source words \emph{$s_i,_1$ ... $s_i,_N$}, where \emph{$t_i,_1$} has the closest embedding to the \emph{$s_i,_1$} embedding in the shared cross-lingual embedding space, typically computed by cosine similarity between the vectors. Afterwards, the retrieved word pairs \emph{N ($s_i,_1$, $t_i,_1$; ...)} are compared to the gold-standard dictionary to measure the model's performance \citep{Ruder2017ASO}. 

The most common reported metric is precision P@\emph{k}, where \emph{k} is the number of the top target words retrieved for one source word. The precision is computed by the following formula:

\begin{equation}
    P = \frac{TP}{(TP + FP)}
\end{equation}

Where \emph{TP} (true positives) is the number of the retrieved word pairs matching the word pairs from the evaluation dataset, and \emph{FP} (false positives) is the number of the retrieved word pairs that do not occur in the evaluation dataset. However, what the papers actually report is HitRatio@\emph{k}, where HitRatio@1 = P@1 and P@\emph{$k_1$} $>$ P@\emph{$k_2$} as long as \emph{$k_1$}$>$\emph{$k_2$} \citep{Conneau2017WordTW}. This is problematic because it overlooks the fact that most source words correspond to multiple target words, and the quantity of these target words varies. Also, it can artificially inflate results above 100\% when \emph{k} exceeds 1. 

Other metrics that could be utilised to measure the models' performance on the BLI task are recall (R) and F1 score. Recall calculates the number of \emph{TP} to the number of \emph{TP} and \emph{FN} (false negatives) combined, where \emph{FN} is the number of word pairs from the evaluation dataset that the model did not retrieve. The recall is defined as follows:

\begin{equation}
    R = \frac{TP}{(TP + FN)}
\end{equation}

The last metric, the F1 score, expresses both metrics, P and R, as one number, and it is determined by the following formula:

\begin{equation}
    F1 = \frac{(2 * P * R)}{(P + R)}
\end{equation}

Recall is an important metric in lexicography, often more relevant than precision since it shows the percentage of the induced words from the evaluation dataset, which are set to be more important than the others. However, the last two mentioned evaluation metrics are scarcely reported in papers. Moreover, there have been some endeavours to advocate for metrics often used in information retrieval to evaluate document ranking systems, such as Mean Average Precision (MAP) \citep{glavas-etal-2019-properly} or Average Precision (AP) \citep{Adjali2022OverviewOT}, but have not been employed in many papers since. We offer an in-depth analysis of the problematic points of the evaluation metrics in Section~\ref{sec:discuss}. In Subsection~\ref{sec:bench}, we outline and discuss the benchmarks and datasets for evaluating models on the BLI task in more detail.

In this survey, we collected recent papers dealing with the BLI task between 2020 and 2023 to map various approaches thoroughly. Depending on how the approaches tackle the polysemy problem, we divided them into two major groups: static and dynamic.

Static cross-lingual embedding models do not change with the context, while dynamic ones take the polysemy of words into account \citep{Wang2020FromST}. For instance, dynamic models distinguish different meanings of the English word \emph{present} based on their contexts (\emph{the present day} or \emph{Christmas present} etc.) by creating different word vector representations for each context in which the word occurs. This property could offer a solution for capturing words’ multiple senses. However, as shown in Section~\ref{sec:dynamic}, current models usually do not utilise this property. Instead, they use only one-word vector representation created by averaging or summing all vector representations for all word senses. Static models operate on a similar basis. They create only one vector representation for all contexts where the word present occurs. 

According to this proposed classification, we summarise the models in Sections~\ref{sec:static} and~\ref{sec:dynamic}. Section~\ref{sec:discuss} shows summary Tables~\ref{tab:1}-~\ref{tab:4} of all methods from Sections~\ref{sec:static} and~\ref{sec:dynamic}.

Even though some of the models might fit into multiple categories, we always choose the category for the concrete model according to its most significant feature. Additionally, we focus on the natural language engineering aspects stated in Section~\ref{sec:treeq} and put them in the context of listed cross-lingual embedding models. 

\subsection{Benchmarks}
\label{sec:bench}

The benchmark methods used for evaluation are frequently MUSE \citep{Conneau2017WordTW, Lample2017UnsupervisedMT}, VecMap \citep{artetxe-etal-2016-learning, artetxe-etal-2017-learning, artetxe2018aaai, artetxe-etal-2018-robust}, and FastText for multilingual alignment \citep{joulin-etal-2018-loss} frameworks \cite[and many others]{glavas-etal-2019-properly, ren-etal-2020-graph, woller-etal-2021-neglect, severini-etal-2022-dont}.

\begin{figure}
  \centering
  \caption{\label{fig:1} The framework of the MUSE model \citep{Conneau2017WordTW}.}  \includegraphics[width=\textwidth,height=\textheight,keepaspectratio]{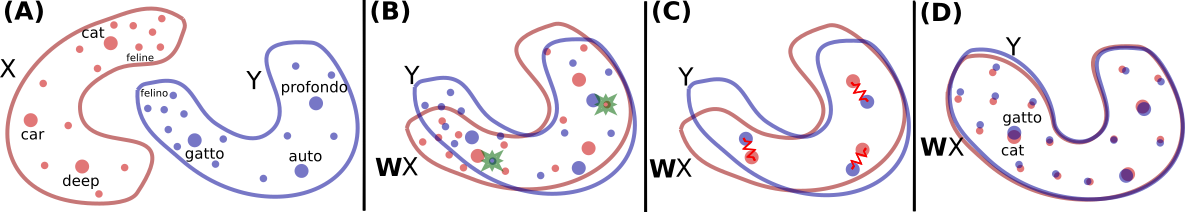}
\end{figure}

MUSE was introduced by \cite{Conneau2017WordTW} and \cite{Lample2017UnsupervisedMT} (two equal articles), and it combines domain-adversarial settings with the Procrustes algorithm. Figure~\ref{fig:1}~\citep{Conneau2017WordTW} illustrates the model's framework. \textbf{(A)} shows two monolingual word embedding distributions, the English, which is marked by \emph{X} in red, and the Italian, which is marked by the \emph{Y} in blue. Firstly, in \textbf{(B)}, the mapping \emph{W} between the target and source space is trained to prevent the discriminator from identifying the origin of an embedding. Then, in \textbf{(C)}, the Procrustes algorithm is iteratively applied to refine the mapping. In the last step \textbf{(D)}, the translations can be extracted from the aligned spaces. For that, a novel method for nearest neighbour search in the cross-lingual space was proposed called Cross-Domain Similarity Local Scaling (CSLS) to avoid the hubness problem~\citep{Conneau2017WordTW}.\footnote{Hubness is an issue observed in high-dimensional space where some points are the nearest neighbours of many other points \citep{Radovanovi2010TimeSeriesCI}.}  

The model MUSE can be trained in a supervised manner using either a seed dictionary or identical strings and in an unsupervised setting. 

MUSE has an open-source GitHub repository\footnote{\href{https://github.com/facebookresearch/MUSE}{https://github.com/facebookresearch/MUSE}} where they published pre-trained cross-lingual embeddings for 30 languages and training and evaluation datasets for 110 languages from and to English. These datasets are widely used as gold-standard dictionaries for evaluating models in the BLI task. They contain 1500 headwords, which is currently a standard for most evaluation datasets \citep{Conneau2017WordTW, glavas-etal-2019-properly, vulic-etal-2019-really, izbicki-2022-aligning}. 

VecMap is a robust self-learning framework proposed by \cite{artetxe-etal-2016-learning, artetxe-etal-2017-learning, artetxe2018aaai, artetxe-etal-2018-robust}, where the small seed lexicon is used to learn embedding mapping. Then, the embedding mapping induces a bilingual dictionary iteratively until a set threshold is reached. In addition, they experimented with seed lexicons of different sizes and numeral seed lexicons. They proved that embedding mapping could be learned with as few as 25 word pairs.

Except for the supervised and semi-supervised training, \cite{artetxe2018aaai} introduced an unsupervised solution that comprises four sequential steps: embedding normalisation, fully unsupervised initialisation that builds an initial dictionary, iterative self-learning procedure that utilises orthogonal mapping, and final refinement step that improves the mapping through symmetric re-weighting iteratively. Both are available in the open-source GitHub repository.\footnote{\href{https://github.com/artetxem/vecmap}{https://github.com/artetxem/vecmap}}

FastText for multilingual alignment was presented by \cite{joulin-etal-2018-loss}. They framed BLI as a retrieval task and adjusted the CSLS retrieval method introduced by \cite{Conneau2017WordTW} by minimising a convex relaxation of the CSLS loss to reduce the hubness problem. Firstly, they applied orthogonal mapping and retrieved the nearest neighbours with the modified CSLS method. Their code is available in the GitHub repository\footnote{\href{https://github.com/facebookresearch/fastText/tree/master/alignment/}{https://github.com/facebookresearch/fastText/tree/master/alignment/}}, and their pre-trained cross-lingual embeddings for more than 44 language pairs are part of the FastText library.\footnote{\href{https://fasttext.cc/docs/en/aligned-vectors.html}{https://fasttext.cc/docs/en/aligned-vectors.html}}  

However, the gold-standard dictionaries from the MUSE repository, i.e., MUSE datasets, face criticism as they might not be a reliable source for evaluation. Based on the study by \cite{kementchedjhieva-etal-2019-lost}, it was found that these dictionaries contain much noise in the form of proper names,  which are not representative enough to mirror the quality of the model. For example, \emph{Barack Obama}, \emph{Skype}, etc. \citep{kementchedjhieva-etal-2019-lost}. Also, incorrect translation equivalents occur in them as they are made automatically. For example, \emph{waves} (plural) - \emph{welle} (singular).\footnote{Example from the MUSE evaluation dataset for English-German.} Therefore, they suggested finding a more trustworthy evaluation method or evaluating with rigorous error analysis. Another similar automatically compiled evaluation dataset is the one used by VecMap authors \citep{Dinu2015ImprovingZL}.

Other datasets utilised by some authors as gold-standard dictionaries are provided by \cite{vulic-etal-2019-really} and \cite{glavas-etal-2019-properly}. Different to the MUSE datasets, both consider various language pairs and combinations (not only to and from English). However, these datasets are generated automatically from the most frequent words or lexical databases. Thus, they do not emphasise the choice of words based on their part of speech or any other aspect. 

An effort to unify the evaluation process and bring high-quality evaluation dictionaries was made by \cite{izbicki-2022-aligning}. This paper addresses the problems of the MUSE datasets stated in \cite{kementchedjhieva-etal-2019-lost}. They manually annotated translations extracted from Wiktionary\footnote{\href{https://en.wiktionary.org/}{https://en.wiktionary.org/}} in 298 languages in combination with English. They focused on a uniformly distributed part of speech in each dataset to make the results as comparable as possible.

As we show in the following sections, many more datasets are used for the evaluation in the BLI task. The evaluation process and datasets are crucial for lexicography as they provide insight into the quality of the models and can answer some of the questions stated in Section~\ref{sec:treeq}. For example, whether authors reflect on how to determine the correct translation of the word, what word form we want the model to generate or what senses of the word to include and mark as correct. Therefore, we factor the evaluation datasets into the survey and list them together with all the natural language engineering aspects in the summary tables~\ref{tab:1}-~\ref{tab:4} in detail. If a method distinguishes itself from the other methods in the evaluation dataset usage, we will also make a note of it within the main part of the survey.  

\section{Static models}
\label{sec:static}

As mentioned above, static cross-lingual embedding models do not depend on the context of the words. Specifically, taking the word present from the previous example, these models create only one vector representation for this word, regardless of its context. 

There are several methods that static models utilise. Namely, joint methods that optimise cross-lingual objectives with monolingual objectives of source and target language at the same time jointly \citep{Ruder2017ASO}, methods that are trained to learn transformation matrix to project monolingual word embeddings of source and target language to the same shared cross-lingual word embedding space or various unsupervised methods that use either adversarial or non-adversarial solutions \citep{Wang2020FromST}.  

In the following subsections, we divide static cross-lingual embedding models into unsupervised (and semi-supervised) and supervised ones and describe the main solutions used.  

\subsection{Unsupervised and semi-supervised models}

Recent research has focused on approaches requiring a tiny or no supervision level. Most models rely on monolingual data and assume that monolingual embedding spaces exhibit similar geometric structures, i.e., are isomorphic, which is not necessarily true. For instance, not every source word has the same number of equivalents in the target language (English article \emph{the} and German \emph{der}, \emph{die}, \emph{das}) \citep{sogaard-etal-2018-limitations}. However, some of the current efforts in the unsupervised methods try to address this issue \citep{mohiuddin-etal-2020-lnmap, wang-etal-2021-multi, Cao2021WordET, nishikawa-etal-2021-data}.

As proven, unsupervised models are still behind their supervised counterparts, and even a small level of supervision (small seed dictionary, identical strings, or numerals) can improve the model’s performance significantly \citep{sogaard-etal-2018-limitations, vulic-etal-2019-really}. Despite this, unsupervised models offer solutions for low-resource languages, as their main objective is not to use any seed dictionary or parallel data.  

This section provides an overview of different approaches unsupervised models utilise. We distinguish several approaches based on research conducted by \cite{ruder-etal-2019-unsupervised}. Specifically, approaches focused on initial seed lexicon induction, adversarial and non-adversarial solutions, and robust approaches consisting of several steps. At the end of the section, we investigate recent efforts which extract translation equivalents through unsupervised machine translation. 

\subsubsection{Methods based on seed lexicon induction}

Methods in this category aim to induce an initial seed lexicon using shared numerals, strings, cognates\footnote{Cognates are words from the same etymological background with similar meaning and orthographic features \citep{kondrak-etal-2003-cognates}.}, named entities, orthographic features, etc. \citep{ruder-etal-2019-unsupervised}. Moreover, authors present various transliteration techniques for languages with different scripts \citep{chakravarthi-etal-2020-bilingual, severini-etal-2020-lmu}. Research in this area suggests that these solutions can noticeably improve the model's results when training in an unsupervised manner \citep{sogaard-etal-2018-limitations}. On the other hand, shared strings may not always convey identical meanings, even among closely related language pairs (e.g., en \emph{also} vs de \emph{also}). 

\cite{ren-etal-2020-graph} used pre-trained monolingual word embeddings to build a graph for each language where vertices represent words, and their mutual information is preserved. Afterwards, they extracted word cliques from the graph using the Bron-Kerbosh algorithm, calculated their embeddings, and mapped them with the \cite{artetxe2018aaai} approach. This built an initial solution for inducing a seed lexicon, which was later used to learn the mapping iteratively with the Procrustes algorithm and refinement procedure. The goal of the graph method was to maintain semantic information of the words and reduce noise from monolingual word embeddings. This is considered beneficial as research suggests that monolingual word embeddings capture more information, and their improvement can lead to better mapping results \citep{artetxe-etal-2018-uncovering}. On the other hand, the authors claim that the overuse of hyper-parameters makes the training unstable. 

\cite{Riley2020UnsupervisedBL} followed \cite{artetxe2018aaai} method and extended it using the orthographic features of different languages. Their method was based on the observation that many languages share borrowed words with the same root. These words often differ in pronunciation or spelling but still preserve many similarities. They continued in the work of \cite{riley-gildea-2018-orthographic}, which proved this observation for languages with the same scripts. In this paper, they extended their solution for languages that do not share scripts. They introduced three techniques: orthographic embedding extension, learned edit distance, and character recurrent neural networks. They also chose different language combinations for the evaluation, such as Polish-Russian and under-resourced Hindi-Bengali, for which the solution has shown to be valuable. However, the authors used pivot dictionaries~\footnote{The pivot dictionary is the outcome of merging two separate dictionaries that share a common language, resulting in the new language pair dictionary.} for the evaluation that were compiled automatically from the MUSE datasets, and their reliability is questionable. 

\citet{chakravarthi-etal-2020-bilingual} focused on low-resource Dravidian languages\footnote{Tamil, Malayalam, Kannada, and Telugu languages} that use different scripts. To align cognates efficiently, they transferred all languages into Latin script. They incorporated \cite{riley-gildea-2018-orthographic}'s approach, in which they substituted Levenshtein distance with the Longest Common Subsequence (LCS) while using \cite{artetxe-etal-2018-robust}'s framework. LCS was utilised for generating morphological variations in previous research \citep{hulden-etal-2014-semi, sorokin-2016-using}, and they showed that LCS is better for determining cognates’ similarity. Unlike the majority of methods, they used their own evaluation datasets compiled from IndoWordNet. Despite the lack of their description and compilation guidelines, it can be regarded as a credible source, given its compilation by lexicographers \citep{pushpak_2010}. 

A similar approach was presented by \cite{severini-etal-2020-lmu} at the BUCC 2020 conference when participating in a shared task. Their focus was to improve the accuracy of low-frequency words and experiment with language pair that does not share an alphabet. They adopted \cite{braune-etal-2018-evaluating}'s approach and utilised the orthographic similarity and RNN transliteration model. Afterwards, they mapped the word embeddings with the VecMap framework \cite{artetxe-etal-2018-robust}. 

In their latter work \citep{severini-etal-2022-dont}, they proposed two ways to easily induce bilingual supervision signals for low-resource, distant languages or languages with different scripts. The first way was to utilise identical strings, which they believed occur in the monolingual data even between distant languages or languages that do not share an alphabet. The second way was to leverage different transliteration and cognate-finding techniques and integrate them into the seed lexicon’s induction. They tested their method on a wide variety of languages, including Arabic, Hindi, Russian, Greek, Persian, Hebrew, Bengali, Korean, Thai, Chinese, Japanese, Kannada, and Tamil (in combination with English). They also provided experiments between Thai, Korean, and Hebrew (in both directions). Although these easily obtainable supervision signals can increase the model's performance significantly, the downside of both of these methods is that orthographically similar words are not always identical in their meanings.

\subsubsection{Adversarial methods}
\label{sec:adv}

Generative adversarial networks (GAN) were proposed by \cite{Goodfellow2014GenerativeAN}, and they are one of the solutions when inducing a bilingual lexicon in an unsupervised way. The framework introduces a min-max two-player game where generative and discriminative models are trained simultaneously while the discriminative model tries to prevent the generative model from making accurate predictions \citep{Goodfellow2014GenerativeAN}. The known drawbacks of these models are that they are hyper-parameter sensitive; some of them exhibit hyper-parameter reliance more than others \citep{Li2021AdversarialTW}, which makes them generally difficult to train.	

The underlying architecture of adversarial models can be illustrated on the MUSE framework example \citep{Conneau2017WordTW}, the most cited method so far \citep{ruder-etal-2019-unsupervised}.

\cite{mohiuddin-etal-2020-lnmap} addressed the isomorphy assumption problem and introduced a new framework, LNMap, that emphasises the importance of non-linear mapping. They suggested that non-linear mapping preserves rich structural information from monolingual word embeddings and requires minimal supervision. This avoids the assumption of isomorphism and enables to focus on low-resource and distant languages. In LNMap, they trained two auto-encoders with non-linear mappers for each language separately. 

Moreover, mapping was limited by two constraints. Firstly, vectors transformed from the source to the target language and back, had to remain close to the original vectors. Secondly, the original input word must be preserved from the back-translated codes.

\cite{Li2021AdversarialTW}'s approach focused on improving accuracy for low-frequency words. They implemented the Wasserstein Generative Adversarial Network. The model consisted of four elements: noise function for adding noises to the source embeddings, auto-encoder encoder (generator) and decoder, and neural network that computed Wasserstein distance. The role of the noise function was to make the embedding space more dispersed. This was followed by their observation that low-frequency words tend to cluster densely. During the evaluation process, apart from using the MUSE dataset, they included also UB-data\footnote{Public dataset extracted from Wikipedia corpora by \cite{zhang-etal-2017-adversarial}.}, and their own dataset constructed from Wikipedia corpora.

\cite{wang-etal-2021-multi} proposed multi-adversarial learning. One adversarial network with multiple discriminators learns different linear mapping for each subspace of word embeddings rather than one linear mapping for the whole space. This was based on an observation that models’ accuracy fluctuates widely for different groups of words gathered in subspaces. The approach aimed to address the isomorphy issue and improve performance for distant language pairs. They claim that the quality of monolingual embeddings and the distribution matching step are the key factors for a high-quality cross-lingual embedding model.

\cite{Bai2022DualityRF} implemented regularisation for a model to enhance correct mapping in both directions. This came from an observation that the model does not map the same word correctly from the source to the target language and back. They added two regularisers to the adversarial model proposed by \cite{Conneau2017WordTW}. Their function was to prevent the mappings from the source to the target language and back from contradicting each other.

\cite{edmiston-etal-2022-domain} tried to address the problem of decreasing accuracy when monolingual corpora domain between languages mismatches. In the previous study by \cite{sogaard-etal-2018-limitations}, it was proven that a mismatch between domains of monolingual corpora used in training caused worse results. \cite{edmiston-etal-2022-domain} claimed this could be solved by using joint pre-training on concatenated multilingual corpora from different domains. They implemented their solution into the MUSE framework. Their method worked mainly for closer language pairs, while it yielded less pronounced improvements for distant languages with different scripts. 

\subsubsection{Non-adversarial methods}

Models in this category attempt to induce the bilingual lexicon in a non-adversarial way, focusing mainly on minor tricks and hyper-parameters such as pre- and post-processing steps. Models often utilise Wasserstein distance, Sinkhorn distance, etc. \citep{ruder-etal-2019-unsupervised}. Non-adversarial solutions offer an alternative to adversarial training that is often expensive, difficult, and unstable. 

On the other hand, the authors report mainly the results for high-frequency words \citep{jawanpuria-etal-2020-geometry, zhao-etal-2020-relaxed, kementchedjhieva-etal-2020-generalizing}, which limits their performance. Moreover, the methods are dependent on the initialization steps, making them sensitive to the noise that monolingual word embeddings contain.   

\cite{jawanpuria-etal-2020-geometry} defined learning of the cross-lingual word embedding mapping as a transformation of the information across domains, known as the domain adaptation task. In their work, they developed a Manifold Based Alignment algorithm that matches covariances from the source to target languages, utilising stochastic matrices' geometry. Moreover, their method maps embeddings in both directions, i.e., from the source to the target language and back. This method observes an improvement on distant language pairs.

\cite{zhao-etal-2020-relaxed} adopted \cite{Grave2018UnsupervisedAO}'s method that minimises the Wasserstein distance for matching monolingual embedding spaces and extends it with the Relaxed Matching Procedure to avoid redundant matchings. Furthermore, they introduced bidirectional optimisation to learn the mapping in both directions (from the source language to the target one and reversely). The results suggest that their method reduces incorrectly matched translation equivalents caused by polysemy or unknown words.

\cite{kementchedjhieva-etal-2020-generalizing} chose an extension of Procrustes Analysis, known as Generalized Procrustes Analysis, to map source and target language monolingual embedding spaces into a third, latent embedding space. They compared both algorithms in their study. As a result, Generalized Procrustes Analysis performs better for synonyms, antonyms, and syntactically related words. As an exemption among researchers in this field, they provided an analysis of the incorrectly aligned word pairs from the linguistic perspective. The revealed problems were with antonyms and morphological variations. They marked it as a mistake if the model induced the word in its inflected form instead of the basic one. Moreover, according to their observations, the model often failed if it aligned a word with a synonymic meaning instead of the word from the evaluation dataset.

Additionally, they mentioned the importance of the BLI task in cross-lingual learning. They focused on mapping English with languages from different language families, concretely, Arabic, German, Spanish, Finnish, and Russian, and on low-resource languages such as Hebrew, Afrikaans, Occitan, Estonian, and Bosnian.

\cite{feng-etal-2022-cross} proposed a novel method that extracts semantic features from monolingual corpora by computing the character distance, aiming to capture similar representations for words with the same meaning in different languages. The method integrates the extracted semantic features using embedding combination and similarity combination while emphasising the initialisation step as key to high-quality alignment. Their approach was based on the assumption that every language possesses similar context-based semantic roles. They focused mainly on improving the performance for low-resource and distant language pairs.

\subsubsection{Robust methods}

Several steps, including iterative refinement and bootstrapping techniques, make methods more robust and allow them to achieve state-of-the-art results. However, some research studies propose that the main advantage of these methods lies in their iterative refinement and stochastic seed dictionary induction, while the actual mapping process itself may be a weaker aspect compared to adversarial solutions \citep{hartmann2019unsup}. 

Representative of this category is \cite{artetxe-etal-2017-learning, artetxe-etal-2018-robust}, with a multi-step framework VecMap in its fully unsupervised and semi-supervised mode, which is also one of the frequently cited benchmarks for many projects in the research of cross-lingual embeddings.  

\cite{karan-etal-2020-classification} proposed CLASSYMAP, a semi-supervised framework with a complex self-learning procedure. They modified the self-learning step with a trained, supervised classifier to assign probability scores to each translation equivalent at each iteration stage. The probability score represents the similarity of the given translation equivalent. Moreover, the classifier integrates various features, such as edit distance, sub-word level similarity, frequencies, pre- and suffixes recognition, etc. They provided an extensive evaluation for twenty-eight language pairs. They presented significant improvements over the benchmark model for rare language combinations.

\cite{Cao2021WordET} extended the VecMap framework by embedding pre-processing through transformation between normalisation and initialisation steps. The transformation consists of rotation and scaling the original embeddings into the same cross-lingual space to increase isomorphism. In their further study, they showed that embedding rotation captures bilingual signal and can serve as a replacement for fully unsupervised methods. 

\cite{Sannigrahi2022IsomorphicCE}'s method introduces a combination of supervised and unsupervised settings. They aimed to leverage the available parallel data of high-resource languages to improve mapping for low-resource languages in the unsupervised scenario, as they believed they could transfer linguistic information from related languages. Their approach consisted of several steps. First, they found related high- and low-resource languages that show a higher level of isomorphism and mapped them in an unsupervised manner. Then, they performed supervised mapping of two high-resource languages that possess enough parallel data (where one of them is a high-resource language from the previous step). In the last step, they aligned the resulting embeddings from the earlier steps together in an unsupervised manner. For unsupervised setting, they used the VecMap framework, and for supervised, they followed \cite{luong-etal-2015-bilingual}'s method. Except for the MUSE datasets, they also used \cite{pavlick-etal-2014-language}'s datasets to evaluate their method.

\cite{Cao2023BilingualWE} focused on addressing the isomorphism problem. Their method operated under the premise that by making the geometric structures of the source and target embeddings more similar, it is possible to improve the alignment and, thus, the translation quality between two languages, especially for those etymologically and typologically distant. Firstly, they introduced the use of Singular Value Decomposition (SVD) to represent the geometric structure of the embeddings. This representation allowed for the quantification and manipulation of isomorphism between embedding spaces. Afterwards, they implemented a robust method through embedding fusion, where features from source embeddings were integrated into target ones and vice versa, using original monolingual embeddings. The embedding fusion consisted of several steps: rotation, joint scaling, vocabulary cut-off, and re-normalisation. In their work, they mainly focused on languages that do not share scripts.     

\subsubsection{BLI task through unsupervised machine translation}

Recent advances in cross-lingual word embedding research recommend not retrieving translation equivalents directly but instead focusing on non-direct solutions, such as using machine translation systems \citep{artetxe-etal-2019-bilingual}. This offers parallel data augmentation, which is particularly beneficial for low-resource languages or uncommon language pairs. However, generating a pseudo-parallel corpus for learning monolingual word embeddings can decrease their quality due to noise \citep{nishikawa-etal-2021-data}.

\cite{nishikawa-etal-2021-data} argued that using an unsupervised machine translation system for cross-lingual embeddings can mitigate the structural differences and retain similarities between two sets of monolingual embeddings. In other words, this system can address the isomorphy problem. In their work, they trained a phrase-based statistical machine translation system on monolingual corpora in an unsupervised manner. Then, they generated a pseudo-parallel corpus by translating the training corpus and concatenating it with the original corpus. Afterwards, they use the pseudo corpus to train monolingual embeddings. In the final step, they incorporated these monolingual embeddings into unsupervised cross-lingual mapping learning. They evaluated the method on dictionaries automatically created through Google Translate\footnote{\href{https://translate.google.com/}{https://translate.google.com/}}.

\subsection{Supervised models}

This subsection introduces the supervised models as a counterpart to the unsupervised ones described in the previous subsection. The supervised scenario often shows better outcomes than the unsupervised one \citep{vulic-etal-2019-really}. However, the required amount of supervision signals is often unrealistic to gather for rare language combinations or low-resource languages.

These methods have been thoroughly researched and surveyed \citep{Ruder2017ASO}. Their main objective is to learn cross-lingual embedding space using a bilingual supervision signal. Bilingual supervision signals can be in the form of word-to-word alignment, sentence-to-sentence alignment, or even document-to-document alignment \citep{Ruder2017ASO}.  

This subsection focuses mainly on methods utilising word-to-word supervision signals (often referred to as bilingual seed lexicon), often of a size of up to 5K word pairs \citep{vulic-korhonen-2016-role}. We outline supervised methods that aspire to address issues connected to a low-resource scenario or improve the projection matrix by utilising various algorithms or machine learning setups. Moreover, we show recent trends concentrating on processing the input monolingual word embeddings to enhance the quality of the resulting cross-lingual embedding space.

\subsubsection{Classification-based methods}

This category presents approaches inspired by the zero-shot learning method \citep{Chang2008ImportanceOS}. The main idea behind zero-shot learning is to train the classifier over labelled data by exploring observable properties. Then, based on the classifier’s observations, it predicts unseen data left out of the training dataset. The advantage is that it could be utilised in the BLI task to infer missing word pairs from the training dataset \citep{artetxe2018aaai}.

Representative of this category is VecMap in its supervised version \citep{artetxe2018aaai}, where the authors adopted a zero-shot learning paradigm where a classifier is trained to predict unseen labels based on training labels, in this case, target words. 

\cite{marchisio-etal-2021-alignment} combined statistical alignment and the zero-shot learning supervised mapping method VecMap \citep{artetxe2018aaai}. They utilised IBM Model 2 \citep{brown-etal-1993-mathematics} to induce an initial seed lexicon on a small parallel corpus. Afterwards, they mapped monolingual embedding via the multi-step VecMap framework in a supervised manner using the previously retrieved seed lexicon. Nonetheless, the authors avoided involving low-frequency words since statistical alignment performs well for frequent occurrences while it tends to fail for less common words.  

In their latter work \citep{marchisio-etal-2021-analysis-euclidean}\footnote{This comparison study does not fit within this category. However, this study is important to mention as it reflects on preserving linguistic information and types of alignments (i.e., one-to-one or many-to-one, etc.).}, they distinguished between viewing word embeddings as a Euclidean space and as weighted graphs with words as nodes and cosine distance as edges. They argued that graphs could preserve rich language information (e.g., syntactic roles, hyponymy/hypernymy, synonymy, etc.). They framed the BLI task in the following two points of view, Euclidean space solved with the Procrustes algorithm and graphs solved by Seeded Graph Matching \cite[SGM]{Fishkind2012SeededGM} and made comparisons between them. Procrustes can pair many-to-one alignments, while SGM is a one-to-one alignment only. Therefore, it does not handle words with multiple meanings. Moreover, their experiments were restricted to one-to-one alignment. 

Similarly, \cite{Marchisio2022BilingualLI} presented a new graph-matching algorithm, GOAT and combined them with the previous method \citep{marchisio-etal-2021-analysis-euclidean}. They believed that GOAT shows better performance when matching non-isomorphic graphs. During the evaluation, they considered only one-to-one alignments.

\subsubsection{Methods based on processing monolingual word embeddings}

In this category, we placed methods whose primary focus is not on learning the transformation matrix or projection itself but on the quality of the monolingual embeddings instead. These methods work with the idea that monolingual input embeddings are an essential component when generating high-quality cross-lingual word embeddings \citep{vulic-etal-2020-improving}. The research suggests that the quality of the cross-lingual embedding models depends strongly on the quality of the monolingual word embeddings and that the methods do not utilise their full potential yet \citep{artetxe-etal-2018-uncovering}. 

\cite{vulic-etal-2020-improving} focused not on learning projection matrix to construct cross-lingual embedding space but on post-processing of the monolingual input embeddings, as these can increase the quality of resulting cross-lingual embeddings. They proposed an unsupervised post-processing method, in which the main objective is to find general linear transformation to adjust first- and second-order similarity. This method’s objective is to capture more linguistic information (such as semantics and words’ similarity). Afterwards, they applied the projection-based supervised framework VecMap to learn the transformation matrix. They published results for diverse languages, among which were some atypical, such as Catalan, Esperanto, Basque, Georgian, Bokmål, etc. Among others, the datasets for the evaluation were also from \cite{dubossarsky-etal-2020-secret}.

\cite{laville-etal-2020-taln} participated in the BLI shared task at the BUCC 2020 conference. They incorporated three methods into the supervised VecMap framework. Firstly, they pre-trained two monolingual word embedding models and used them for supervised training in the VecMap framework. In the second method, they concatenated both monolingual embedding models and repeated the process with supervised VecMap training. Based on the analysis, the third method assumed that low-frequency word pairs are often graphically identical. Therefore, they applied the perfect cognates matching approach for matching similarly spelt strings. To increase the performance, they combined the last two methods. Additionally, they critically analysed seed lexicons for training. The analysis revealed many mistakes; for instance, the seed lexicon contained words from other languages, morphological incoherence between translation equivalents, spelling errors, proper names, etc. 

\cite{eder-etal-2021-anchor} focused on training monolingual word embeddings for the high-resource language and utilised them to initialise the low-resource language monolingual space. Initially, they trained monolingual word embedding models for the source language with rich resources. Afterwards, they started to shape the vector space of the target low-resource language by replacing the target words from the bilingual seed lexicon with vector representations of the source words. Their aim was to solve the problem when one of the languages lacked monolingual resources. They evaluated their method with the help of MUSE datasets and datasets provided by \cite{michel-etal-2020-exploring} on English-German, English-Hiligaynon, and English-Macedonian language pairs. 

\cite{ormazabal-etal-2021-beyond} method aimed to learn monolingual embeddings for the source and target language together. They argued that separate training causes mismatches in the embedding structure. Firstly, they trained monolingual embeddings for the target language. Afterwards, they constrained the source embeddings to be close to the target ones using seed words as anchor points as an extended version of the monolingual word embedding model. Finally, they improved the mapping iteratively via a robust self-learning procedure.

\cite{ZHOU2022116194} proposed a novel method for learning cross-lingual word embeddings with auxiliary topic models. They chose auxiliary topic models as one of the ways to capture semantic information. They focused on improving monolingual word embeddings and seed lexicon as they considered them the key factors for high-quality cross-lingual embedding models. In the first approach, they iteratively refined topic and embedding models to obtain monolingual word embeddings and incorporated them into projection-based cross-lingual embedding mapping. The second approach concerned retrieving bilingual seed lexicon from bilingual topic models based on bilingual semantic similarities. The seed lexicon was induced and used for mapping in a supervised mode and iteratively generated under unsupervised settings. The authors also provided results for Spanish-Italian and German-Italian, among others.

\subsubsection{Projection-focused methods}

The main objective of methods in this category is to improve or find a suitable projection matrix to produce better cross-lingual word embeddings. Methods are often focused on mitigating the issue with non-isomorphic monolingual word embedding spaces. Therefore, the distant language pairs whose monolingual word embedding graphs are not isomorphic benefit primarily from these solutions. On the other hand, these methods are strictly focused on the mathematical side of the BLI task, rarely reflecting on the linguistic side (with exceptions, e.g., \cite{Tian2022RAPOAA}). 

\cite{glavas-vulic-2020-non} introduced the framework INSTAMAP, which differs from learning a linear projection matrix in using non-parametric training. In their method, two steps are iteratively repeated: firstly, the Kabsch algorithm is applied to learn globally optimal rotation, and secondly, each word’s translation vector is computed by averaging the translation vectors of the nearest neighbours. Their solution improves the performance of distant language pairs with a low degree of monolingual embedding spaces’ isomorphism. The results were published for 28 various language pairs (not only to and from English).

\cite{jp-etal-2020-bucc2020} presented their method for learning cross-lingual word embeddings to solve the shared task at the BUCC 2020 conference. They utilised deep neural networks (DNN) for transfer learning and pre-trained monolingual word embeddings. The authors conducted a reverse search to determine the correct word translation in the original embedding space, using the transfer-learned embedding as a reference.

Initially, the source embeddings were fed to the DNN to produce target embeddings that were compared to the original target embedding in the next phase. The closest embedding was considered as the given source word’s translation equivalent. They aimed to mimic the language semantics and transfer it from the target to the source language. The authors saw the importance of the BLI task in machine translation. They proposed a new algorithm instead of using gold-standard dictionaries. This algorithm follows an observation that better cross-lingual spaces have higher similarity scores and compares the original vector to the learned vector.  

\cite{Tian2022RAPOAA} proposed a novel framework, RAPO (Ranking-based model with Adaptive Personalized Offsets), that involves three steps. First, a contextual vector is computed for each word by averaging the vectors of the neighbour words. The motivation was to capture more semantic information. Then, the Householder projection is applied to align the word embeddings into shared space. They chose this projection as it could encourage the isomorphism of the word embedding spaces. The last step consists of ranking the induced translation equivalent candidates to help differentiate between the correctly and incorrectly aligned word pairs.

\subsubsection{Methods focused on low-resource scenario}

Current state-of-the-art methods perform well for languages with a large amount of data available. Still, they produce poor results or fail entirely for low-resource, distant, or morphologically rich languages \citep{sogaard-etal-2018-limitations}. Moreover, most models' accuracy on non-frequent words is still very low \citep{huang-etal-2020-improving}, or they did not report on these at all \cite[etc.]{jawanpuria-etal-2020-geometry, zhao-etal-2020-relaxed, kementchedjhieva-etal-2020-generalizing}. 

One of the alternatives is unsupervised models. However, they are under criticism for their assumption of isomorphism of the monolingual spaces, which has been proven incorrect \citep{sogaard-etal-2018-limitations}. Hence, this category introduces methods that aim to research and address the main issues connected to either low-resource languages or low-frequency words in a supervised setting. Recent advances in dynamic models have also demonstrated efforts to enhance performance for low-resource languages (see Section~\ref{sec:stat_dyn}). 

The main limitation of these methods is that they often present results for low-resource languages paired with English \citep{michel-etal-2020-exploring, hakimi-parizi-cook-2021-evaluating} (with exceptions, e.g., \cite{woller-etal-2021-neglect, Resiandi2023NeuralNB}), which in fact have sufficient resources in a real scenario (e.g., MUSE datasets).  

\cite{huang-etal-2020-improving} focused on low-frequency words and factors that affect the decreasing accuracy of these words. As the source and target word frequencies could differ significantly, they introduced monolingual lexicon induction to mitigate these frequency discrepancies between the two languages. They trained monolingual word embeddings from two corpora for one language. They mapped these two sets of embeddings using orthogonal transformation and 500K seed words, which were split into 50 frequency groups. Afterwards, they observed that diminishing margin between cosine similarities when retrieving nearest neighbours and hubness issue play an important role in accuracy for low-frequency words. They suggested two algorithms to improve two above-stated issues: Hinge Loss for Learning Transformation for enlarging the margin and Hubless Nearest Neighbour Search. 

\cite{michel-etal-2020-exploring} followed \cite{braune-etal-2018-evaluating}'s research and studied how corpora and seed lexicon size influence the model's performance for low-resource language Hiligaynon. Firstly, they set up a low resource scenario for close and distant language pairs and mapped monolingual word embeddings using \cite{Mikolov2013EfficientEO}'s
method. Furthermore, they experimented with seed lexicons of various sizes and observed the impact. The conclusion was that the method fails for both languages if trained using small corpora. However, using the same size corpora and smaller seed lexicon can produce better results than training conducted with a larger seed lexicon. They published results for the English-Hiligaynon language pair and compared them with English-German. The English-Hiligaynon language pair was evaluated on a manually constructed dictionary by a native speaker. For English-German, they used \cite{braune-etal-2018-evaluating}'s datasets. 

Additionally, in Hiligaynon, senses often consist of multiple words, such as plurals or comparative word forms, etc. Therefore, they pointed out that constructing a dictionary for Hiligaynon with monolingual and cross-lingual word embedding models is problematic as these do not handle multi-word expressions. Moreover, they emphasised that the polysemy problem should be solved by monolingual word embeddings having different vectors for each word’s sense. 

\cite{hakimi-parizi-cook-2020-joint} adopted the method presented in the paper by \cite{duong-etal-2016-learning} and extended it by using sub-word information in the word embedding training. In their approach \citep{duong-etal-2016-learning}, they utilised \cite{Mikolov2013EfficientEO}'s model trained on monolingual corpora with bilingual seed lexicon. \cite{hakimi-parizi-cook-2020-joint} incorporated the sub-word information in the cross-lingual embedding joint training inspired by \cite{bojanowski-etal-2017-enriching}'s method. They offered this solution for low-resource and morphologically rich languages as a way how to handle out-of-vocabulary words. They claimed that many methods fail for low-resource and morphologically rich languages because less frequent word forms are missing their vector representation in the monolingual word embeddings. These vector representations could also be formed for missing words when enriching the method with sub-word information. They tested their method on high-resource languages while simulating a low-resource scenario. 

Later, \cite{hakimi-parizi-cook-2021-evaluating} evaluated this extended method \citep{hakimi-parizi-cook-2020-joint} on even lower-resource languages, namely, Afrikaans, Albanian, Azerbaijani, Bengali, Bosnian, Croatian, Estonian, Greek, Hebrew, Hindi, Hungarian, Turkish. In both articles, they used MUSE datasets for the evaluation, except for Azerbaijani, where they tested their method with \cite{anastasopoulos-neubig-2020-cross}’s dataset. 

\cite{woller-etal-2021-neglect} presented a novel approach that aims to improve results in the BLI task for low-resource language Occitan. They adopted \cite{Wang2019CrosslingualAV} joint-align method to map Occitan with related languages (French, Spanish, and Catalan) that possess higher resources. Then, the learned cross-lingual embedding space is used for supervised training with the MUSE framework \citep{Conneau2017WordTW} to create one shared multi-lingual space with the English language. They believed that utilising linguistic information from related languages could produce better results. For the evaluation, they used several resources, such as MUSE datasets, \cite{kementchedjhieva-etal-2020-generalizing}'s datasets, the Occitan website\footnote{http://www.occitania.online.fr/}, and freelang.\footnote{https://www.freelang.com/}

\cite{Resiandi2023NeuralNB} focused on endangered low-resource Indonesian and Minangkabau languages, aiming to preserve them by leveraging neural networks to extract phonetic transformation rules. Their technique employed character-level embeddings and Bi-LSTM in a sequence-to-sequence model for deep phonetic and morphological analysis. The model processed Indonesian text character-by-character, creating a context-rich representation, which the LSTM decoder then used to generate Minangkabau text. Afterwards, they compared their approach to a rule-based baseline. Importantly, they automatically compiled evaluation and training data that were validated by native speakers, making it exceptional compared to the other approaches.

\section{Dynamic models}
\label{sec:dynamic}

Dynamic models could offer the solution to effectively address the polysemy problem and the problem of capturing multiple senses of the word. Dynamic word embedding representations or contextualised vectors are not fixed but change with the context \citep{Wang2020FromST}. This means they create different vector representations for the same word occurring in different contexts. 

However, many models do not work with different word vectors for one word separately. Instead, they average or sum all these word vector representations into one, representing all word’s senses together \citep{schuster-etal-2019-cross}, making them similar to the static models. For the dynamic models to be successful, they need to generate multiple meanings and multiple vectors for those meanings. Moreover, some researchers suggest that static models are still exceeding the performance of dynamic models, and they are a more appropriate solution for the BLI task \citep{zhang-etal-2021-combining, li-etal-2022-improving}. Despite recent advances showing promising results \citep{li-etal-2023-bilingual}, a notable challenge remains: the current evaluation datasets are tailored for assessing static models while they do not reflect various contexts or senses of the words. Therefore, these datasets do not assess the full potential of the dynamic models, and the results do not provide an accurate picture of the models' performance.     

Another advantage is that some dynamic models, for instance, BERT \citep{devlin-etal-2019-bert}, have an in-built tokeniser that divides a word into smaller tokens and utilises sub-word information. This might be a solution for tackling out-of-vocabulary words \citep{hakimi-parizi-cook-2020-joint}.   

Most of the current contextualised cross-lingual word embedding models extract contextual word embeddings from sequence-to-sequence transformer-based attention models that were first introduced by \cite{Vaswani2017AttentionIA}. Following the idea presented by \cite{schuster-etal-2019-cross}, they collect multiple context vectors for a word type and summarise them to get an average anchor point. Afterwards, these anchor points are utilised for further mapping. Alternatively, other methods focus on non-direct translation equivalents retrieval through a machine translation system \citep{shi-etal-2021-bilingual} or define the problem as density matching \citep{Zhao2022ConstrainedDM}. Some endeavours attempt to combine the strengths of static and dynamic models \cite[etc.]{zhang-etal-2021-combining}.

On top of that, the release of ChatGPT has spawned an immense number of research articles incorporating large language models (LLMs) into various NLP tasks, including BLI. The trend is currently moving from the static models, promoting the dynamic ones with a pioneering rigorous study of the utilisation and application of LLMs for the BLI task \citep{li-etal-2023-bilingual}. 

This section provides an overview of the current dynamic models and describes the advances in this field mentioned above. Namely, it distinguishes attention-based methods, transformer-based methods through unsupervised machine translation, and combined methods and introduces a recent study on LLMs.

\subsection{Attention-based models}

In many NLP fields, transformer-based attention models first introduced by \cite{Vaswani2017AttentionIA} are currently trending. Transformers are sequence-to-sequence models with an attention mechanism that decides which words within the context are important at each step. These models gained popularity as they can achieve state-of-the-art results in many NLP tasks. However, their extensive training requires a huge amount of data that is not always available, especially for low-resource languages.  

Some transformer models are pre-trained for multiple languages, such as a multi-BERT \citep{devlin-etal-2019-bert} multilingual model trained on concatenated corpora consisting of 104 languages. Also, multilingual models XLM and XLM-R \citep{Lample2019CrosslingualLM} were inspired by the BERT model and pre-trained for one hundred languages or similar models as Rosita \citep{mulcaire-etal-2019-polyglot}, mBART \citep{liu-etal-2020-multilingual-denoising}, CRISS \citep{Tran2020CrosslingualRF} and many more. 

However, these above-mentioned multilingual transformer models do not have aligned embedding spaces between the languages and are not pre-trained for the BLI task. Therefore, one solution is to capture multiple context vectors for a word type and conflate these vectors into the average anchor points. Then, these anchor points could be leveraged for mapping the monolingual embedding spaces to each other. Such cross-lingual embedding spaces are often utilised for zero-shot dependency parsing task \citep{schuster-etal-2019-cross, wang-etal-2019-cross}. However, this section discusses dynamic models that focus mainly on the BLI task.

Another drawback of these methods is that most utilise parallel corpus, making it difficult for low-resource or rare language pairs. The size of the parallel corpus is usually small \citep{Xu2021CrossLingualBC, wada-etal-2021-learning}, and the focus is on cross-lingual embedding models. Therefore, we list these methods among comparable data-based methods.

On top of that, attention-based contextual cross-lingual embedding models utilise different types of training, for example, joint training \citep{wada-etal-2021-learning}, training with causal language modelling objective \citep{zhou-etal-2022-prix}, or adjusting the spaces to be isotropic and isometric \citep{Xu2021CrossLingualBC}. Next, we outline these methods.  

\cite{Xu2021CrossLingualBC} focused on increasing isomorphism of the monolingual embedding spaces and multi-sense words. They argued that contextual word embeddings show a higher level of isomorphism and can tackle the multi-sense word problem. They extracted multiple contextual vectors for a word type by feeding a parallel corpus into the BERT transformer model \citep{devlin-etal-2019-bert}. Afterwards, they computed the mean vector for each word type and called these vectors type-level embeddings. They split type-level embeddings into sense-level embeddings, where each embedding represents one sense of the word. Obtained embeddings were aligned by enhancing the shared space to be isotropic and isometric. Additionally, the authors considered different types of alignment, such as one-to-one, many-to-one, and many-to-many.   

\cite{wada-etal-2021-learning} implemented a bi-directional encoder-decoder LSTM model with attention to training contextualised word embeddings. Then, they performed joint training to learn shared embedding space using a small parallel corpus where linguistic information among the languages can be transferred. Their method aimed to preserve sub-word information, that is, orthographic features and the context of the words. In their experiments, they focused mainly on low-resource endangered languages, such as Yongning Na, Shipibo-Konibo, and Griko. Also, they provided an evaluation for high-resource languages, concretely English, Japanese, Inuktitut, German, and French. They tested the method with datasets from \cite{maguino-valencia-etal-2018-wordnet} and \cite{Michaud2018NaD}. 

\cite{zhou-etal-2022-prix} proposed the knowledge-based cross-lingual model Prix-LM. They adopted the concept of the model XLM-R \citep{conneau-etal-2020-unsupervised}. They pre-trained it on DBpedia’s multilingual knowledge base \citep{Lehmann2015DBpediaA} with a causal language modelling objective. DBpedia is a knowledge base that consists of the descriptions of entities’ relations and cross-lingual links between these entities. The result was the model that maps knowledge from multiple languages into the same unified space. This could help to enrich low-resource languages through close high-resource language pairs. For the BLI task, they extracted translation equivalents using the supervised VecMap framework. They tested their method with datasets provided by \cite{glavas-etal-2019-properly} on 17 languages, among which they included Telugu, Lao, and Marathi. 

\subsection{Transformer-based methods with an unsupervised machine translation}

Followed by the idea introduced by \cite{artetxe-etal-2019-bilingual} that current advances should focus on non-direct translation equivalents retrieval (i.e., through machine translation system), there has been an attempt to combine unsupervised machine translation and contextualised word embeddings. Such effort was proposed by \cite{shi-etal-2021-bilingual}, and this category discusses their method.

\cite{shi-etal-2021-bilingual} was inspired by \cite{artetxe-etal-2019-bilingual}'s approach and incorporated unsupervised machine translation into their method. In contrast, they replaced the statistical machine translation model with transformer models CRISS and mBART to produce higher-quality translations. They combined the unsupervised bitext mining, which includes unsupervised machine translation and bitext retrieval and unsupervised word alignment that induces bilingual lexicon. Their method was based on the criticism that many models unrealistically assume similar embeddings for one-word translations. In addition, they trained the model in an unsupervised and semi-supervised manner. They considered only one-to-one alignments for the unsupervised mode and semi-supervised, one-to-one, and many-to-one. 

Moreover, they aimed to improve the results for low-frequency words. They evaluated their method with the help of MUSE and BUCC conference (2020) datasets. Also, they took \cite{kementchedjhieva-etal-2020-generalizing}'s research into account and performed a manual evaluation.

This method is unique in several ways. Firstly, unlike the majority of papers, the authors report F1 scores, providing a complete picture of the model's performance. Moreover, they reflect on multiple drawbacks that have been pointed out in previous studies and many authors omitted in their papers, such as inaccurate evaluation datasets (addressed by a manual evaluation) and polysemy problems. 

\subsection{Density-based methods}

The density matching approach was proposed as opposed to the large data requiring multi-lingual contextual word embedding models. Following the observation that contextualised models are not properly trained to utilise the data sufficiently, \cite{Zhao2022ConstrainedDM} proposed a density-based method involving an adversarial solution. Similarly to the methods in Section \ref{sec:adv}, such solutions are computationally demanding and difficult to train. This category describes these efforts. 

\cite{Zhao2022ConstrainedDM}, motivated to decrease the amount of data for training and therefore make the method more available for the low-resource scenario, they incorporated density matching techniques into the contextualised word embedding training for the BLI task. Firstly, they trained the model in a supervised scenario which showed two problems: ineffective data usage and a lack of criteria to correctly select a better model. Afterwards, they suggested aligning subspaces of the multilingual space as probability distributions. They implemented the Wasserstein generative adversarial network and Earth Movers Distance and trained the model in an unsupervised mode. To address the lack of validation criteria problem, they computed the semantic \cite{Lample2017UnsupervisedMT} and the structural criterion \citep{dubossarsky-etal-2020-secret}. \footnote{Semantic criterion averages cosine similarities of the induced translation equivalents. Structural criterion computes the difference between induced word pairs and gold-standard word pairs.} 

In the evaluation part, they used automatically extracted word pairs from the parallel corpora. Additionally, they used an automatic gold-standard dataset and a real dictionary containing noises. However, they did not reveal the exact source of their evaluation datasets, making their method even less comparable to the other ones.

\subsection{Combined static and dynamic representations}
\label{sec:stat_dyn}

Although most of the recent research has been concentrated either on static or dynamic methods, some methods were proposed that connect properties of both these representations. The research on static and dynamic models suggests that static models outperform the dynamic ones and are more apt for the BLI task \citep{zhang-etal-2021-combining, li-etal-2022-improving}. However, the combination of these two models can enhance the performance and leverage their strengths \citep{zhang-etal-2021-combining}. We describe these attempts in this category.

\cite{zhang-etal-2021-combining} introduced a method that combines static and dynamic word embedding representations. For static embeddings, they followed the VecMap framework \citep{artetxe2018aaai, artetxe-etal-2018-robust}. The dynamic representations were inspired by \cite{schuster-etal-2019-cross} anchor method that collects contextual embeddings for a word type and averages them. Firstly, the unified space is built with the spring network that is trained to enhance the position of the static embeddings by using contextual embeddings. Additionally, similarity interpolation was performed between the unified embeddings and contextual embeddings. 

\cite{li-etal-2022-improving} combined a cross-lingual embedding model with a pre-trained multilingual language model. Their method consisted of two steps; first, they trained the VecMap framework for static word embeddings. Second, they fine-tuned the multilingual language model mBERT with correctly and incorrectly aligned word pairs extracted from the previous step as examples. Lastly, they combined word embeddings from both steps and utilised them for the BLI task. Although they involved a dynamic model, they induced static word embeddings. They considered various language combinations among the following languages: Croatian, English, Finnish, French, German, Russian, Italian, Turkish, Basque, Catalan, Bulgarian, Estonian, Hebrew, and Hungarian.          

\cite{vulicetal2022} studied rich lexical knowledge preserved in the multilingual language models\footnote{mBERT and XLM-R \citep{conneau-etal-2020-unsupervised}} and sentence encoders\footnote{LaBSE \citep{feng-etal-2022-cross}, xMPNET \citep{reimers-gurevych-2019-sentence}}, and how they could gain from this knowledge the maximum for the cross-lingual transfer learning. They interpolated these methods with a static cross-lingual embedding approach using the VecMap framework.

\cite{el-mekki-etal-2023-promap} introduced ProMap, a novel approach leveraging the power of prompting pre-trained multilingual and multi-dialectal language models through a technique known as padded prompting. This approach aims to overcome the limitations posed by using subword tokens in these models, which can complicate the word translation process between languages. ProMap consists of two primary variants: ProMapG (Generation) and ProMapS (Selection). The former is aimed at low-resource languages, generating translations directly from a pre-trained language model in few-shot scenarios. Conversely, the latter uses an existing static word embedding mapping method to select the correct translation from multiple candidates. 

On top of that, the authors provided a rigorous empirical evaluation across multiple language pairs, including various dialects of Arabic. They evaluated their method on automatically compiled datasets from \cite{glavas-etal-2019-properly} and \cite{erdmann-etal-2018-addressing} and exploited manually compiled data with translation guidelines from \cite{bouamor-etal-2018-madar}.

This paper also falls into the category "Methods focused on the low-resource scenario", which has been a focus of more recent publications connecting low-resource languages and the strengths of dynamic models. For example, \cite{artemova-plank-2023-low} leveraged the large language models for bitext mining followed by employing machine translation aligners aiming to align Bavarian and Alemannic dialects with standard German. Similarly, \cite{Bafna2023ASM} exploited model BERT pre-trained for high-resource language to make predictions for closely related low-resource language.  

\subsection{Recent study}

After ChatGPT was released, the number of papers published incorporating the LLMs into different NLP tasks skyrocketed, and BLI is no exception. Most recent trends show the transition from the traditional static approaches towards dynamic ones. The more rigorous study that discusses using and employing the LLMs in the BLI task was conducted by \cite{li-etal-2023-bilingual}.

\cite{li-etal-2023-bilingual} explored the application of multilingual LLMs in the BLI task and assessed their effectiveness across three main strategies: zero-shot prompting for unsupervised BLI, few-shot in-context prompting without model fine-tuning, and BLI-oriented fine-tuning of smaller multilingual LLMs. The study utilised 18 open-source text-to-text multilingual LLMs, ranging from 0.3B to 13B parameters and provided detailed evaluation, including ablation studies. The evaluation covered a variety of language pairs in both directions.

However, as the article mentioned, the LLMs are unavailable for as many languages as the static models. Moreover, the authors underscored that the evaluation datasets employed in their study, originally designed for assessing static models, fall short in addressing polysemy. Consequently, these datasets may not fully capture the nuanced capabilities of dynamic models in handling the complexities of language.

\section{Discussion on the BLI task}
\label{sec:discuss}

This section outlines all discussed methods in summary Tables~\ref{tab:1},~\ref{tab:2},~\ref{tab:3},~\ref{tab:3.1.},~\ref{tab:4}. Tables~\ref{tab:1}-~\ref{tab:3.1.} show static models divided into supervised and unsupervised (and semi-supervised). Table~\ref{tab:4} displays dynamic models. 

Each table shows the tracked aspects stated in Section~\ref{sec:treeq}, whether the method addresses or discusses morphology (Morph.), senses/semantics/polysemy (Senses), and multi-word expressions (MWE). Moreover, it outlines languages (Languages) and datasets used in the evaluation process (Eval./ Evaluation). Finally, it displays the models’ purpose, if mentioned in the article (Purp./Purpose).

\begin{table}
\caption{Summary of the important natural language engineering aspects of the static unsupervised methods I. MT = machine translation, ST = standalone task.}
\centering
    \begin{tabular}{lccccl}
    \toprule
    \multicolumn{6}{c}{{\bf Static - Unsupervised I.}} \\  \midrule
    {\bf Article}& \multicolumn{1}{c}{{\bf Morph.}}& \multicolumn{1}{c}{{\bf Senses}}& {\bf MWE}& {\bf Purp.}& {\bf Eval.} \\  
    {\bf Languages}& \\ \midrule
    \multicolumn{6}{l}{\bf Methods based on seed lexicon induction} \\
    \midrule
    \citet{ren-etal-2020-graph}& $\checkmark$& $\checkmark$& $$-$$ & MT & MUSE \\ 
    \multicolumn{3}{l}{ fr, de, es, it, ru, zh, fi, pl, tr $\rightleftarrows$  en}& \\
     \citet{Riley2020UnsupervisedBL}& $$-$$& $$-$$& $$-$$& MT& MUSE, pivot \\  
     en-ru, pl-ru, hi-bn \\
    \citet{chakravarthi-etal-2020-bilingual}& $\checkmark$& $$-$$& $$-$$& $$-$$& \multirow{2}{2cm}{IndoWordNet, VecMap} \\ 
    Dravidian languages \\
    \citet{severini-etal-2020-lmu}& $$-$$& $$-$$& $$-$$& MT& MUSE \\ 
    de-en, en-de, ru-en, en-ru \\
    \citet{severini-etal-2022-dont}& $$-$$& $$-$$& $$-$$& $$-$$& \multirow{2}{2.5cm}{MUSE, \citet{vulic-etal-2019-really}} \\  
    \multicolumn{6}{l}{ar, hi, ru, el, fa, he, bn, ko, th, zh, ja, kn, ta $\rightleftarrows$ en, th $\rightleftarrows$ ko $\rightleftarrows$ he} \\
    \midrule
    \multicolumn{6}{l}{\bf Adversarial methods} \\
    \midrule
    \citet{mohiuddin-etal-2020-lnmap}& $$-$$& $\checkmark$& $$-$$& $$-$$& \multirow{2}{2cm}{MUSE, VecMap} \\ 
    \multicolumn{5}{l}{ms, fi, et, tr, el, fa, he, ta, bn, hi, es, de, it, ar, ru $\rightleftarrows$ en} \\
    \citet{Li2021AdversarialTW}& $$-$$& $$-$$& $$-$$& $$-$$& \multirow{2}{2.7cm}{MUSE, UB-data, own data} \\ 
    \multicolumn{5}{l}{hu, af, bg, ru, hr, de, fr, es, zh, tr, it $\rightleftarrows$ en}\\
    \citet{wang-etal-2021-multi}& $$-$$& $$-$$& $$-$$& $$-$$& MUSE \\ 
    \multicolumn{5}{l}{de, es, fr, it, ja, ru, tr, zh $\rightleftarrows$ en} \\
    \citet{Bai2022DualityRF}& $$-$$& $\checkmark$& $$-$$& $$-$$& \multirow{2}{2cm}{MUSE, VecMap} \\  
    it, de, fi, es $\rightleftarrows$ en \\
    \citet{edmiston-etal-2022-domain}& $$-$$& $$-$$& $$-$$& ST & MUSE \\ 
    de, fr, es, it, ru, fa $\rightleftarrows$ en \\
    \bottomrule
    \end{tabular}
    \label{tab:1}
\end{table}

\begin{table}
\caption{Summary of the important natural language engineering aspects of the static unsupervised methods II. SET = sentence translation, UMT = unsupervised machine translation, CIR = cross-lingual information retrieval, CDP = cross-lingual dependency parsing, GT = google Translation.}
    \begin{tabular}{lccccl}
    \toprule
    \multicolumn{6}{c}{{\bf Static - Unsupervised II.}} \\  \midrule
    {\bf Article}& \multicolumn{1}{c}{{\bf Morph.}}& \multicolumn{1}{c}{{\bf Senses}}& {\bf MWE}& {\bf Purpose}& {\bf Eval.}\\
    {\bf Languages}
    & \\ \midrule
    \multicolumn{5}{l}{\bf Non-adversarial methods} \\
    \midrule
    \citet{jawanpuria-etal-2020-geometry}& $$-$$& $\checkmark$& $$-$$& $$-$$& MUSE \\
    \multicolumn{5}{l}{de, es, fr, it, pt, bg, cs, da, el, fi, hu, nl, pl, ru, ar, hi, tr $\rightleftarrows$ en} \\
    \citet{zhao-etal-2020-relaxed}& $$-$$& $\checkmark$& $$-$$& \multirow{2}{1.5cm}{SET, UMT, CIR}& MUSE \\
    es, fr, de, it, ru $\rightleftarrows$ en \\
    \citet{kementchedjhieva-etal-2020-generalizing}& $\checkmark$& $\checkmark$& $$-$$& \multirow{2}{1.5cm}{MT, cross-lingual learning}& \multirow{2}{2.5cm}{MUSE, VecMap, \citet{Dinu2015ImprovingZL}} \\
    \multicolumn{5}{l}{ar, de, es, fi, ru, he, bs, af, oc, et $\rightleftarrows$ en} \\ \\
    \citet{feng-etal-2022-cross}& $$-$$& $\checkmark$& $$-$$& \multirow{2}{1.5cm}{UMT, CDP, CIR}& MUSE \\
    vi, th, ja, zh $\rightleftarrows$ en \\
    \midrule
    \multicolumn{4}{l}{\bf Robust methods} \\
    \midrule
    \citet{karan-etal-2020-classification}& $\checkmark$& $$-$$ & $$-$$& $$-$$& \multirow{2}{2cm}{\citet{glavas-etal-2019-properly}} \\ 
    \multicolumn{5}{l}{tr-hr, de-tr, tr-fi, tr-ru, fi-hr, de-hr, de-ru, etc.} \\
    \citet{Cao2021WordET}& $$-$$& $$-$$& $$-$$& $$-$$& MUSE \\
    \multicolumn{5}{l}{es, de, fr, ru, ja, zh, vi, th $\rightleftarrows$ en} \\
    \citet{Sannigrahi2022IsomorphicCE}& $$-$$& $\checkmark$& $$-$$& ST& \multirow{2}{2.5cm}{MUSE, \citet{pavlick-etal-2014-language}} \\
    \multicolumn{5}{l}{hi, ne, gu, fi, et, hu, it, ro $\rightleftarrows$ en} \\
    \citet{Cao2023BilingualWE}& $$-$$& $$-$$& $$-$$& \multirow{2}{2cm}{MT, transfer learning}& MUSE \\
    \multicolumn{5}{l}{ru, zh, ja, vi, th, es, de, fr $\rightleftarrows$ en} \\
    \midrule
    \multicolumn{4}{l}{\bf BLI task through UMT} \\
    \midrule
    \citet{nishikawa-etal-2021-data}& $$-$$& $$-$$& $$-$$& $$-$$& \multirow{2}{2cm}{Dictionaries through GT} \\ 
    fr, de, ja $\rightleftarrows$ en \\
    \bottomrule
    \end{tabular}
     \label{tab:2}
\end{table}

\begin{table}
 \caption{Summary of the important natural language engineering aspects of the static supervised methods I.}
 \centering
    \begin{tabular}{lccccl}
    \toprule
    \multicolumn{6}{c}{{\bf Static - Supervised I.}} \\ 
    \midrule
    {\bf Article}& \multicolumn{1}{c}{{\bf Morph.}}& \multicolumn{1}{c}{{\bf Senses}}& {\bf MWE}& {\bf Purp.}& {\bf Eval.}\\  
    {\bf Languages}& \\ \midrule
    \multicolumn{4}{l}{\bf Classif.-based methods} \\
    \midrule
    \citet{marchisio-etal-2021-alignment}& $$-$$& $$-$$& $$-$$& MT, CIR & VecMap \\
    de, es, it, fi $\rightleftarrows$ en \\
    \citet{marchisio-etal-2021-analysis-euclidean}& $$-$$& $\checkmark$& $$-$$& MT, CIR & MUSE \\
    de, ru $\rightleftarrows$ en \\
    \citet{Marchisio2022BilingualLI}& $$-$$& $$-$$& $$-$$& MT, CIR& MUSE \\
    \multicolumn{6}{l}{vi, id, ms, ta, ja, zh, et, mk, bs, ru, de, bn, fa, fr, es, pt, it $\rightleftarrows$ en} \\
    \midrule
    \multicolumn{4}{l}{\bf Methods based on processing MWEs} \\
    \midrule
    \citet{vulic-etal-2020-improving}& $$-$$& $\checkmark$& $$-$$& ST &\multirow{3}{3cm}{\citet{vulic-etal-2019-really}, \citet{dubossarsky-etal-2020-secret}, 
    \citet{glavas-etal-2019-properly}} \\ \\
    \\
    \multicolumn{6}{l}{bg, ca, et, eo, eu, fi, he, hu, id, ka, ko, lt, no, th, tr} \\
    \citet{laville-etal-2020-taln}& $$-$$& $$-$$& $$-$$& $$-$$& MUSE \\
    en, es, de, fr, ru \\
    \citet{eder-etal-2021-anchor}& $$-$$& $$-$$& $$-$$& $$-$$& \multirow{2}{3cm}{MUSE, \citet{michel-etal-2020-exploring}} \\
    de, mk, hil $\rightleftarrows$ en \\
    \citet{ormazabal-etal-2021-beyond}& $$-$$& $$-$$& $$-$$& $$-$$& MUSE \\
    \multicolumn{6}{l}{de, es, fr, fi, ru, zh $\rightleftarrows$ en} \\
    \citet{ZHOU2022116194}& $$-$$& $\checkmark$& $$-$$& $$-$$& MUSE \\
    \multicolumn{6}{l}{it, es, fr, de $\rightleftarrows$ en, es, it $\rightleftarrows$ de} \\
    \bottomrule
    \end{tabular}
\label{tab:3}
\end{table}

\begin{table}
\caption{Summary of the important natural language engineering aspects of the static supervised methods II. SP = semantic parsing, DC = document classification.}
\centering
    \begin{tabular}{lccccl}
    \toprule
    \multicolumn{6}{c}{{\bf Static - Supervised II.}} \\ 
    \midrule
    {\bf Article}& \multicolumn{1}{c}{{\bf Morph.}}& \multicolumn{1}{c}{{\bf Senses}}& {\bf MWE}& {\bf Purp.}& {\bf Eval.}\\  
    {\bf Languages}& \\ \midrule
    \multicolumn{5}{l}{\bf Projection-focused methods} \\
    \midrule
    \citet{glavas-vulic-2020-non}& $$-$$& $$-$$& $$-$$& $$-$$& \multirow{2}{2cm}{\citet{glavas-etal-2019-properly}} \\
    \multicolumn{6}{l}{en, de, fr, it, ru, hr, tr, fi} \\
    \citet{jp-etal-2020-bucc2020}& $$-$$& $\checkmark$& $$-$$& MT& \multirow{2}{2cm}{Own algorithm} \\
    de, ta $\rightleftarrows$ en \\
    \citet{Tian2022RAPOAA}& $$-$$& $\checkmark$& $$-$$& 
    \multirow{2}{1.7cm}{MT, SP, DC}& MUSE \\
    \multicolumn{6}{l}{es, fr, it, ru, zh, fa, tr, he, ar, et $\rightleftarrows$ en} \\
    \midrule
    \multicolumn{5}{l}{\bf Methods focused on low-resource scenario} \\
    \midrule
    \citet{huang-etal-2020-improving}& $$-$$& $$-$$& $$-$$& $$-$$& MUSE \\
    \multicolumn{6}{l}{es, fr, it, ru, zh, fa, tr, he, ar, et $\rightleftarrows$ en} \\
    \citet{michel-etal-2020-exploring}& $$-$$& $$-$$& $$-$$& $$-$$& \multirow{2}{3cm}{\citet{braune-etal-2018-evaluating}, manual hil-en} \\
    hil, de $\rightleftarrows$ en \\
    \citet{hakimi-parizi-cook-2020-joint}& $\checkmark$& $$-$$& $$-$$& $$-$$& MUSE \\
    \multicolumn{6}{l}{zh, nl, fr, de, it, ja, ru, es $\rightleftarrows$ en} \\
    \citet{hakimi-parizi-cook-2021-evaluating}& $\checkmark$& $$-$$& $$-$$& $$-$$& \multirow{2}{3.5cm}{MUSE, \citet{anastasopoulos-neubig-2020-cross}} \\
    \multicolumn{6}{l}{af, alb, ae, bn, bs, hr, et, el, hz, hi, hu, tr $\rightleftarrows$ en} \\
    \citet{woller-etal-2021-neglect}& $$-$$& $$-$$& $$-$$& ST &\multirow{3}{3cm}{MUSE, Occitan web, \citet{kementchedjhieva-etal-2020-generalizing}, freelang} \\
    fr, es, ca, oc, en \\
    \\
    \citet{Resiandi2023NeuralNB}& $\checkmark$& $$-$$& $$-$$& $$-$$& \multirow{2}{3.5cm}{auto-compiled datasets verified by natives} \\
    \multicolumn{6}{l}{Indonesian $\rightleftarrows$ Minangkabau} \\
    \bottomrule
    \end{tabular}
     \label{tab:3.1.}
\end{table}

\begin{table}
  \caption{Summary of the important natural language engineering aspects of the dynamic methods. POS.TAG = part of speech tagging, SG = serious games, LL = language learning.}
  \centering
    \begin{tabular}{lccccl}
    \toprule
    \multicolumn{6}{c}{{\bf Dynamic}} \\
    \midrule
    {\bf Article} & \multicolumn{1}{c}{{\bf Morph.}} & \multicolumn{1}{c}{{\bf Senses}} & {\bf MWE} & {\bf Purpose} & {\bf Evaluation}\\  
    {\bf Languages}
    & \\ \midrule
    \multicolumn{5}{l}{\bf Attention-based models} \\
    \midrule
    \citet{Xu2021CrossLingualBC}& $$-$$& $\checkmark$& $$-$$& $$-$$ & MUSE \\
    de, nl, ar $\rightleftarrows$ en \\
    \citet{wada-etal-2021-learning}& $$-$$& $\checkmark$& $$-$$& $$-$$ & \multirow{2}{3.5cm}{\citet{maguino-valencia-etal-2018-wordnet, Michaud2018NaD}} \\
    \multicolumn{5}{l}{nru, shp, grk, en, ja, de, fr, iu} \\
    \citet{zhou-etal-2022-prix}& $$-$$& $\checkmark$& $$-$$& $$-$$ & \citet{glavas-etal-2019-properly} \\
    \multicolumn{5}{l}{en, de, es, it, fr, ru, fi, et, hu, tr, ko, ja, zh, th, te, lo, mr} \\
    \midrule
    \multicolumn{5}{l}{\bf Transformer-based methods with the UMT} \\
    \midrule
    \citet{shi-etal-2021-bilingual}& $$-$$& $\checkmark$& $$-$$& $$-$$ & MUSE, BUCC (2020) \\
    en, de, fr, es, ru, zh \\
    \midrule
    \multicolumn{5}{l}{\bf Density-based methods} \\
    \midrule
    \citet{Zhao2022ConstrainedDM}& $$-$$& $\checkmark$& $$-$$& $$-$$ & \multirow{2}{3.5cm}{Automatic and manual dictionaries} \\
    \multicolumn{5}{l}{de, cz, lt, fi, ru, tr $\rightleftarrows$ en} \\
    \midrule
    \multicolumn{5}{l}{\bf Combined static and dynamic representations} \\
    \midrule
    \citet{zhang-etal-2021-combining}& $$-$$& $\checkmark$& $$-$$& \multirow{2}{1.5cm}{POS.TAG, parsing, MT} & MUSE \\ 
    \multicolumn{5}{l}{es, ar, zh, de, fr $\rightleftarrows$ en} \\
    \citet{li-etal-2022-improving}& $$-$$& $\checkmark$& $$-$$& $$-$$ & \multirow{2}{3.5cm}{\citet{vulic-etal-2019-really, glavas-etal-2019-properly}} \\
    \multicolumn{5}{l}{hr, en, fi, fr, de, it, ru, tr, bg, ca, eu, et, he, hu} \\
    \citet{vulicetal2022}& $$-$$& $\checkmark$& $$-$$& $$-$$ & \multirow{2}{3.5cm}{\citet{vulic-etal-2019-really, glavas-etal-2019-properly}} \\
    bg, ca, et, he, ka \\
    \citet{el-mekki-etal-2023-promap}& $\checkmark$& $\checkmark$& $$-$$& \multirow{3}{1.5cm}{MT, CIR, LL, SG} & \multirow{3}{3.5cm}{\citet{glavas-etal-2019-properly, erdmann-etal-2018-addressing, bouamor-etal-2018-madar}} \\ \\
    de, fr, it, ru, tr, Arabic dialects \\
    \citet{artemova-plank-2023-low}& $$-$$& $\checkmark$& $$-$$& \multirow{2}{1.5cm}{transfer learning} & \multirow{2}{3.5cm}{manual eval. and dataset construction} \\ 
    \multicolumn{5}{l}{Alemannic, Bavarian $\rightleftarrows$ de} \\
    \citet{Bafna2023ASM}& $$-$$& $\checkmark$& $$-$$& \multirow{2}{2cm}{LL, SP, UMT} & \multirow{2}{2.5cm}{\cite{ojha-etal-2020-findings}} \\ 
    \multicolumn{5}{l}{Indic languages} \\
    \midrule
    \multicolumn{5}{l}{\bf Recent study} \\
    \midrule
    \citet{li-etal-2023-bilingual}& $$-$$& $\checkmark$& $$-$$& \multirow{2}{2cm}{MT, transfer learning} & \multirow{2}{3cm}{\citet{vulic-etal-2019-really, glavas-etal-2019-properly}} \\ 
    \multicolumn{5}{l}{en, de, fi, fr, hr, it, ru, tr, bg, ca, hu} \\
    \bottomrule
    \end{tabular}
      \label{tab:4}
\end{table}

Additionally, this section provides a discussion on listed methods and approaches in the previous sections and frames them into the context of natural language engineering aspects in Section~\ref{sec:treeq}. It explores individual methods, demonstrating their effectiveness in addressing aspects highlighted in Section~\ref{sec:treeq}, and it identifies specific NLP scenarios where employing these methods proves to be particularly beneficial. For example, when dealing with morphologically rich languages, it is important to use the methods and evaluation datasets addressing morphology, as they lead to more precise and accurate outcomes. On the other hand, in some limited cases of NLP applications, such as identifying names for analytic language pairs (e.g., English, Chinese), morphology does not play as pivotal a role.

From an extensive list of the research papers in Tables~\ref{tab:1}-~\ref{tab:4} and a comprehensive overview of the methods in Sections~\ref{sec:static} and~\ref{sec:dynamic}, it is clear that many efforts have been made in the research of the BLI task (and many other articles that did not appear in this survey). We can also expect this task to enjoy a great interest in the future.

Authors of the articles dealing with the BLI task offer a wide variety of methods trying to address several problems occurring during the BLI task, such as better seed lexicon induction \cite[etc.]{ren-etal-2020-graph, severini-etal-2020-lmu}, mitigating the isomorphism problem \cite[etc.]{mohiuddin-etal-2020-lnmap, Cao2021WordET, glavas-vulic-2020-non}, improving monolingual word embeddings used in training \cite[etc.]{vulic-etal-2020-improving, eder-etal-2021-anchor}, hubness issue \cite[etc.]{huang-etal-2020-improving}, enhance the results for low-resource or distant language pairs \cite[etc.]{michel-etal-2020-exploring, woller-etal-2021-neglect}, and many others. 

When involving a lexicography context, we can observe minor efforts to resolve some issues. Firstly, the morphological variations have been challenging in the BLI task \citep{sogaard-etal-2018-limitations, artetxe-etal-2018-robust, yang-etal-2019-maam}. Some methods offered to incorporate sub-word information (words’ stems) to generate translations for less common word forms \cite[etc.]{chakravarthi-etal-2020-bilingual, hakimi-parizi-cook-2020-joint, hakimi-parizi-cook-2021-evaluating, karan-etal-2020-classification}. \cite{ren-etal-2020-graph} observed an improvement for the morphologically rich languages during the evaluation. Moreover, \cite{Resiandi2023NeuralNB} implemented character-level embeddings to capture morphological variations. From the earlier efforts, \cite{yang-etal-2019-maam} introduced a morphology-aware alignment that ensures corresponding word forms align together. 

The only reflection on whether to punish the model when finding various word forms instead of basic word forms was expressed in \cite{kementchedjhieva-etal-2019-lost, kementchedjhieva-etal-2020-generalizing}. The authors considered morphological variations as a mistake and preferred basic word forms. Additionally, based on criticism by \cite{kementchedjhieva-etal-2019-lost}, the MUSE datasets contain morphologically mismatched word pairs. Therefore, we can assume that the articles that used the MUSE datasets for the evaluation do not consider this aspect. 

Another important aspect was the senses and polysemy of the words. Some methods endeavour to utilise dynamic models for capturing different meanings of the word throughout its occurrences in various contexts \cite[etc.]{zhou-etal-2022-prix, shi-etal-2021-bilingual, Zhao2022ConstrainedDM}. Some explicitly claim to use one-to-many or many-to-many alignments \citep{Xu2021CrossLingualBC, shi-etal-2021-bilingual}. 

However, it is not clear yet how to leverage the advantages of the dynamic models for word-to-word mapping \citep{vulicetal2022}. Therefore, many dynamic models average the embeddings of the multiple words’ context \cite[etc.]{zhang-etal-2021-combining, Li2021AdversarialTW}, making them similar to the static models that do not solve polysemy. A similar ‘averaging’ approach appeared in \cite{Tian2022RAPOAA} with static word embeddings. On top of that, all the dynamic models use the evaluation datasets intended for static models, failing to evaluate the polysemy handling ability accurately. 

In the static models, we can observe some efforts to preserve high-quality semantic information by grouping semantically related words into separate graphs \citep{ren-etal-2020-graph, marchisio-etal-2021-analysis-euclidean}, non-linear mapping \citep{mohiuddin-etal-2020-lnmap}, classifying words based on their semantic roles \citep{feng-etal-2022-cross}, leveraging linguistic information from close languages \citep{michel-etal-2020-exploring, Sannigrahi2022IsomorphicCE}, post-processing monolingual word embedding used in training \citep{vulic-etal-2020-improving}, and topic-modelling \citep{ZHOU2022116194}. 

On top of that, some articles focused on mitigating the issue of incorrectly aligned word pairs (also due to the polysemy) by incorporating reverse translations \cite[etc.]{jawanpuria-etal-2020-geometry, zhao-etal-2020-relaxed, Bai2022DualityRF}. Although this solution may work for high-frequency words and basic vocabulary \footnote{For instance, authors reported results for the first 20K occurrences \citep{jawanpuria-etal-2020-geometry, zhao-etal-2020-relaxed}.}, it fails for more complex words due to language asymmetry and ambiguity as word A = word B does not always mean that word B = word A (e.g., en \emph{to run} - fr \emph{courir}, \emph{diriger}, \emph{gérer}), and it can lead to significant loss of information, i.e., recall declines.   

Despite all these efforts to improve semantics, none of the articles reflects on how to evaluate polysemic words and how to deal with multiple senses or multiple target words. Some methods even allow for only one-to-one alignments \citep{marchisio-etal-2021-analysis-euclidean}, some evaluate their methods considering only one induced equivalent \cite[etc.]{Riley2020UnsupervisedBL, severini-etal-2020-lmu, ormazabal-etal-2021-beyond}, and some even build their evaluation datasets allowing only one target word for each source word \citep{glavas-etal-2019-properly}. However, one error analysis of the different lexical-semantic relationships was carried out by \cite{kementchedjhieva-etal-2020-generalizing}.

As for the third aspect, none of the articles dedicated their research to multi-word expressions. However, we can find reflection on multi-word expressions in \cite{michel-etal-2020-exploring}, where authors saw word-to-word translations as problematic. Therefore, the cross-lingual embedding models are still unable to handle them, and employing NMT systems and LLMs remains a more effective approach for this task.

The variety of languages the methods were evaluated across is very diverse. We can find high- (English, German, French, Chinese, etc.) and low-resource languages (Estonian, Japanese, Hiligaynon, Tamil, etc.), languages from different language families (Italian, Finnish, Vietnamese, Malay, Dravidian, etc.), languages with different alphabets (Russian, Arabic, Korean, etc.), morphologically rich languages (Hungarian, Turkish, Czech, etc.), or even dialects (Bavarian, Minangkabau, Indic dialects, etc.).

The main trend is to provide an evaluation with English as either source or target language \cite[etc.]{ren-etal-2020-graph, mohiuddin-etal-2020-lnmap, hakimi-parizi-cook-2020-joint, Li2021AdversarialTW, Bai2022DualityRF, feng-etal-2022-cross}. However, we can find exceptions allowing for rather rare language combinations \cite[etc.]{Riley2020UnsupervisedBL, karan-etal-2020-classification, vulic-etal-2020-improving, li-etal-2022-improving}. Some methods offer multilingual mapping into one shared space \cite[etc.]{woller-etal-2021-neglect}.   

When tracking the purpose of the methods, many articles did not mention further utilisation at all \cite[etc.]{laville-etal-2020-taln, Li2021AdversarialTW, wang-etal-2021-multi, severini-etal-2022-dont} or saw BLI as a standalone task \cite[etc.]{vulic-etal-2020-improving, edmiston-etal-2022-domain, Sannigrahi2022IsomorphicCE}. If the purpose was stated, it was restricted to other NLP tasks, such as machine translation \cite[etc.]{Riley2020UnsupervisedBL, jp-etal-2020-bucc2020}, cross-lingual information retrieval \cite[etc.]{zhao-etal-2020-relaxed, marchisio-etal-2021-alignment, marchisio-etal-2021-analysis-euclidean, Marchisio2022BilingualLI}, dependency parsing \cite[etc.]{zhang-etal-2021-combining, feng-etal-2022-cross}, document classification \citep{Tian2022RAPOAA}, or transfer learning \citep{artemova-plank-2023-low, li-etal-2023-bilingual}, etc. The exception was language learning mentioned in some articles \citep{el-mekki-etal-2023-promap, Bafna2023ASM}. The lexicography field was not included at all. 

Finally, the evaluation dataset is the last aspect. Given Tables~\ref{tab:1}-~\ref{tab:4}, the process of evaluation of the methods is not standardised and varies from model to model, hindering our ability to compare the results and their accurate quality.

The MUSE datasets enjoy big popularity and are widely used among researchers \cite[etc.]{ren-etal-2020-graph, Riley2020UnsupervisedBL, jawanpuria-etal-2020-geometry, zhao-etal-2020-relaxed, Xu2021CrossLingualBC, zhang-etal-2021-combining, edmiston-etal-2022-domain, Tian2022RAPOAA}, despite their criticism by \cite{kementchedjhieva-etal-2019-lost}. Some of the efforts combine MUSE datasets with other resources \cite[etc.]{Li2021AdversarialTW, shi-etal-2021-bilingual, severini-etal-2022-dont, Bai2022DualityRF}. Recent advances prefer datasets created by \cite{glavas-etal-2019-properly} and \cite{vulic-etal-2019-really} \cite[etc.]{li-etal-2022-improving, vulicetal2022} as they support various language combinations. 

Importantly, neither these evaluation datasets nor the methods reflect on translation and how to determine the correct translation equivalent, what senses to include or how to handle inflected word forms. If these factors are included, they are mentioned in the manual error analysis \cite[etc.]{kementchedjhieva-etal-2020-generalizing, shi-etal-2021-bilingual}.

These datasets are mostly automatically generated (MUSE datasets, \cite{glavas-etal-2019-properly}; \cite{vulic-etal-2019-really}; \cite{Dinu2015ImprovingZL}; etc.) and do not consider part of speech distribution or other aspects of the choice of words. Additionally, we have not observed using manually created datasets from \cite{izbicki-2022-aligning} so far.

In conclusion, some of the authors report significant progress, especially for low-resource languages \cite[etc.]{chakravarthi-etal-2020-bilingual, mohiuddin-etal-2020-lnmap, feng-etal-2022-cross} or some dynamic models \citep{Xu2021CrossLingualBC, zhou-etal-2022-prix}. However, as also Table~\ref{tab:results} confirms, the improvements over the benchmark models stated in the majority of the papers are very minor for some language pairs and models, staying within the margin of 5 \% \cite[etc.]{ren-etal-2020-graph, vulic-etal-2020-improving, zhao-etal-2020-relaxed, Cao2021WordET, marchisio-etal-2021-alignment, Bai2022DualityRF, ZHOU2022116194}. This can stem from manifold reasons. Firstly, the authors use problematic evaluation data that do not follow natural language engineering aspects. Secondly, they try to optimise their new solutions and models according to these problematic data, which prevents progress. Also, due to the incoherent evaluation procedures, such as various evaluation datasets, publicly unavailable evaluation datasets \citep{Zhao2022ConstrainedDM} or even novel evaluation algorithms that skip evaluation datasets and have not been adopted by more authors \citep{jp-etal-2020-bucc2020}, it is difficult to correctly interpret the results and compare the outcomes of the models. 

On top of that, the main reported metric is precision \cite[etc.]{Riley2020UnsupervisedBL, wang-etal-2021-multi, ormazabal-etal-2021-beyond, Xu2021CrossLingualBC, Tian2022RAPOAA, li-etal-2022-improving} (actually HitRatio@\emph{k}, see Section~\ref{sec:comp}), whereas recall or F1 scores are hardly presented (rare attempt to report on F1 scores appeared in \cite{severini-etal-2020-lmu} and \cite{shi-etal-2021-bilingual}). This means that although the model has significant gains in precision, its recall can decrease rapidly, especially for models using back-translations \citep{denisova-rychly-2023}. Additionally, some minor efforts advocate for using MAP \citep{glavas-etal-2019-properly} or AP \citep{Adjali2022OverviewOT} scores. However, a review of the literature reveals an absence of subsequent studies that persist in utilizing these metrics.

\begin{table}
\caption{The comparison of the benchmark and authors' results evaluated with the MUSE datasets.}
  \centering
    \begin{tabular}{c|cc}
    \toprule
    Language Pair & Model & Result (\%) \\
    \midrule
    \multirow{4}{1cm}{en-de} & \cite{Conneau2017WordTW} (benchmark) & 74.0 \\
          & \cite{ren-etal-2020-graph} & 75.5 \\
          & \cite{zhao-etal-2020-relaxed} & \textbf{76.2} \\
          & \cite{marchisio-etal-2021-alignment} & 51.7 \\
          & \cite{Bai2022DualityRF} & 70.0 \\
          & \cite{ZHOU2022116194} & 72.28 \\
    \midrule
    \multirow{4}{1cm}{en-fr} & \cite{joulin-etal-2018-loss} (benchmark) & 83.3\\
          & \cite{ren-etal-2020-graph} & 82.9 \\
          & \cite{zhao-etal-2020-relaxed} & 83.0 \\
          & \cite{ZHOU2022116194} & \textbf{83.59} \\
    \midrule
    \multirow{4}{1cm}{en-fi} & \cite{artetxe-etal-2018-robust} (benchmark) & 32.4 \\
          & \cite{vulic-etal-2020-improving} & \textbf{37.4} \\
          & \cite{marchisio-etal-2021-alignment} & 37.1 \\
          & \cite{Bai2022DualityRF} & 22.2 \\
    \bottomrule
    \end{tabular}
  \label{tab:results}
  \end{table}

\section{Conclusion and future work}
\label{sec:conc}

In this article, we have divided the methods for finding translation equivalents into parallel- and comparable-data-based with two subcategories, statistical- and neural-network-based. We have focused mainly on neural-network-based methods utilising comparable data, and we have reasoned our decision. We have collected papers from 2020 to 2023 dealing with the BLI task, classified them based on their most significant feature, and thoroughly described them. We have focused on investigating whether these methods deal with natural language engineering aspects of extracting translation equivalents, such as identifying correct translation equivalents and handling multi-word expressions, polysemy, and inflected word forms.

Moreover, we have tracked languages used for the evaluation, methods’ purposes, and evaluation datasets. Finally, we have discussed the methods from the lexicographic point of view. This survey enables the identification of those methods that consider natural language engineering aspects and implement them, and it offers valuable material for research in the NLP field.

Overall, a great number of studies have been done on this topic, utilising various approaches and solving problems from different points of view. Despite this, the cross-lingual models still need to be properly researched from the lexicographic point of view to gain better results in the NLP field. Many authors aim at plain translation equivalent retrieval in the word-to-word format, and their primary focus is often on method improvement, viewing the issue from an NLP side while neglecting the language context. 

Nevertheless, the cross-lingual embedding models aspire to be a valuable contribution to the NLP field due to their availability for all languages and language pairs and the possibility of using text-balanced corpora simulating real-language scenarios. Given these reasons, they could also serve as supplementary data for the widely used parallel data.

These models still possess drawbacks and challenges, such as morphology variance, handling low-resource and distant languages, words’ senses and polysemy, multi-word expressions, and inconsistent evaluation methods that do not take the linguistic side into consideration, which offer room for further improvements. As for future research directions, we encourage the authors to focus on these limitations and incorporate our lexicographic observations and natural language engineering aspects we pointed out in this survey into their research. 

We emphasise the need for a unified, more transparent, and better methodology for evaluation datasets' compilation and evaluation process itself that reflects the aspects stated in Section~\ref{sec:treeq}. Such methodology is vital for monitoring the progress of the methods accurately, and it makes models comparable. Moreover, the dynamic cross-lingual embedding models should be investigated more to capture the polysemy more precisely and separate evaluation datasets handling the polysemy of the words should be compiled for the dynamic models. Another possible research direction is to examine the possibility of including the multi-expressions in cross-lingual models and evaluation processes. Also, we highlight the importance of reporting the results for various metrics apart from precision and evaluating the models on language pairs that do not include English.     

On top of that, the cross-lingual embedding models have recently drawn a lot of attention, and we expect more and more research to be conducted in this area; therefore, we await improvements in the limitations listed above. 

\bibliographystyle{unsrtnat}
\bibliography{references}

\begin{thebibliography}{148}
\providecommand{\natexlab}[1]{#1}
\providecommand{\url}[1]{\texttt{#1}}
\expandafter\ifx\csname urlstyle\endcsname\relax
  \providecommand{\doi}[1]{doi: #1}\else
  \providecommand{\doi}{doi: \begingroup \urlstyle{rm}\Url}\fi

\bibitem[Conneau et~al.(2017)Conneau, Lample, Ranzato, Denoyer, and J'egou]{Conneau2017WordTW}
Alexis Conneau, Guillaume Lample, Marc'Aurelio Ranzato, Ludovic Denoyer, and Herv'e J'egou.
\newblock Word translation without parallel data.
\newblock \emph{ArXiv}, abs/1710.04087, 2017.
\newblock \doi{/10.48550/arXiv.1710.04087}.

\bibitem[Artetxe et~al.(2017{\natexlab{a}})Artetxe, Labaka, Agirre, and Cho]{Artetxe2018UnsupervisedNM}
Mikel Artetxe, Gorka Labaka, Eneko Agirre, and Kyunghyun Cho.
\newblock Unsupervised neural machine translation.
\newblock \emph{ArXiv}, abs/1710.11041, 2017{\natexlab{a}}.
\newblock \doi{/10.48550/arXiv.1710.11041}.

\bibitem[Artetxe et~al.(2019{\natexlab{a}})Artetxe, Labaka, and Agirre]{artetxe-etal-2019-effective}
Mikel Artetxe, Gorka Labaka, and Eneko Agirre.
\newblock An effective approach to unsupervised machine translation.
\newblock In \emph{Proceedings of the 57th Annual Meeting of the Association for Computational Linguistics}, pages 194--203. Association for Computational Linguistics, 2019{\natexlab{a}}.
\newblock \doi{10.18653/v1/P19-1019}.

\bibitem[Duan et~al.(2020)Duan, Ji, Jia, Tan, Zhang, Chen, Luo, and Zhang]{duan-etal-2020-bilingual}
Xiangyu Duan, Baijun Ji, Hao Jia, Min Tan, Min Zhang, Boxing Chen, Weihua Luo, and Yue Zhang.
\newblock Bilingual dictionary based neural machine translation without using parallel sentences.
\newblock In \emph{Proceedings of the 58th Annual Meeting of the Association for Computational Linguistics}, pages 1570--1579. Association for Computational Linguistics, 2020.
\newblock \doi{10.18653/v1/2020.acl-main.143}.

\bibitem[Lavrenko et~al.(2002)Lavrenko, Choquette, and Croft]{lavrenko2002}
Victor Lavrenko, Martin Choquette, and W.~Bruce Croft.
\newblock Cross-lingual relevance models.
\newblock In \emph{Proceedings of the 25th annual international {ACM SIGIR} conference on Research and development in information retrieval}, pages 175--182. ACM, 2002.

\bibitem[Xiao and Guo(2014)]{Xiao2014DistributedWR}
Min Xiao and Yuhong Guo.
\newblock Distributed word representation learning for cross-lingual dependency parsing.
\newblock In \emph{Proceedings of the Eighteenth Conference on Computational Natural Language Learning}, pages 119--129. Association for Computational Linguistics, 2014.
\newblock \doi{10.3115/v1/W14-1613}.

\bibitem[Guo et~al.(2015)Guo, Che, Yarowsky, Wang, and Liu]{guo2015parsing}
Jiang Guo, Wanxiang Che, David Yarowsky, Haifeng Wang, and Ting Liu.
\newblock Cross-lingual dependency parsing based on distributed representations.
\newblock In \emph{Proceedings of the 53rd Annual Meeting of the Association for Computational Linguistics and the 7th International Joint Conference on Natural Language Processing (Volume 1: Long Papers)}, pages 1234--1244. Association for Computational Linguistics, 2015.
\newblock \doi{10.3115/v1/P15-1119}.

\bibitem[Klementiev et~al.(2012)Klementiev, Titov, and Bhattarai]{klem2012}
Alexandre Klementiev, Ivan Titov, and Binod Bhattarai.
\newblock Inducing crosslingual distributed representations of words.
\newblock In \emph{Proceedings of {COLING} 2012}, pages 1459--1474. The COLING 2012 Organizing Committee, 2012.
\newblock URL \url{https://aclanthology.org/C12-1089}.

\bibitem[Yuan et~al.(2020)Yuan, Zhang, Van~Durme, Findlater, and Boyd-Graber]{yuan-etal-2020-interactive}
Michelle Yuan, Mozhi Zhang, Benjamin Van~Durme, Leah Findlater, and Jordan Boyd-Graber.
\newblock Interactive refinement of cross-lingual word embeddings.
\newblock In \emph{Proceedings of the 2020 Conference on Empirical Methods in Natural Language Processing (EMNLP)}, pages 5984--5996. Association for Computational Linguistics, 2020.
\newblock \doi{10.18653/v1/2020.emnlp-main.482}.

\bibitem[Denisová and Rychlý(2023)]{denisova-rychly-2023}
Michaela Denisová and Pavel Rychlý.
\newblock Evaluation of the cross-lingual embedding models from the lexicographic perspective.
\newblock In \emph{Proceedings of the eLex 2023 conference}, pages 1--18. Lexical Computing CZ, s.r.o., 2023.

\bibitem[Fontenelle(1997)]{fontenelle1997}
Thierry Fontenelle.
\newblock \emph{Turning a bilingual dictionary into a lexical-semantic database}.
\newblock Lexicographica, 1997.

\bibitem[Atkins and Rundell(2008)]{atkins2008oxford}
Beryl T.~Sue Atkins and Michael Rundell.
\newblock \emph{The Oxford Guide to Practical Lexicography}.
\newblock Oxford University Press, New York, 2008.

\bibitem[Jakubíček et~al.(2021)Jakubíček, Kovář, and Rychlý]{jakubivcek2021million}
Miloš Jakubíček, Vojtěch Kovář, and Pavel Rychlý.
\newblock Million-click dictionary: {T}ools and methods for automatic dictionary drafting and post-editing.
\newblock In \emph{The {XIX} {EURALEX} International Congress: {L}exicography for inclusion}, pages 65--67, 2021.

\bibitem[Zhang et~al.(2017)Zhang, Liu, Luan, and Sun]{zhang-etal-2017-adversarial}
Meng Zhang, Yang Liu, Huanbo Luan, and Maosong Sun.
\newblock Adversarial training for unsupervised bilingual lexicon induction.
\newblock In \emph{Proceedings of the 55th Annual Meeting of the Association for Computational Linguistics (Volume 1: Long Papers)}, pages 1959--1970. Association for Computational Linguistics, 2017.
\newblock \doi{10.18653/v1/P17-1179}.

\bibitem[Jawanpuria et~al.(2020)Jawanpuria, Meghwanshi, and Mishra]{jawanpuria-etal-2020-geometry}
Pratik Jawanpuria, Mayank Meghwanshi, and Bamdev Mishra.
\newblock Geometry-aware domain adaptation for unsupervised alignment of word embeddings.
\newblock In \emph{Proceedings of the 58th Annual Meeting of the Association for Computational Linguistics}, pages 3052--3058. Association for Computational Linguistics, 2020.
\newblock \doi{10.18653/v1/2020.acl-main.276}.

\bibitem[Cao and Zhao(2021)]{Cao2021WordET}
Hailong Cao and Tiejun Zhao.
\newblock Word embedding transformation for robust unsupervised bilingual lexicon induction.
\newblock \emph{ArXiv}, abs/2105.12297, 2021.
\newblock \doi{/10.48550/arXiv.2105.12297}.

\bibitem[Wang et~al.(2021)Wang, Henderson, and Merlo]{wang-etal-2021-multi}
Haozhou Wang, James Henderson, and Paola Merlo.
\newblock Multi-adversarial learning for cross-lingual word embeddings.
\newblock In \emph{Proceedings of the 2021 Conference of the North American Chapter of the Association for Computational Linguistics: Human Language Technologies}, pages 463--472. Association for Computational Linguistics, 2021.
\newblock \doi{10.18653/v1/2021.naacl-main.39}.

\bibitem[Severini et~al.(2022)Severini, Hangya, Jalili~Sabet, Fraser, and Sch{\"u}tze]{severini-etal-2022-dont}
Silvia Severini, Viktor Hangya, Masoud Jalili~Sabet, Alexander Fraser, and Hinrich Sch{\"u}tze.
\newblock Don{'}t forget cheap training signals before building unsupervised bilingual word embeddings.
\newblock In \emph{Proceedings of the BUCC Workshop within LREC 2022}, pages 15--22. European Language Resources Association, 2022.

\bibitem[Mikolov et~al.(2013{\natexlab{a}})Mikolov, Le, and Sutskever]{Mikolov2013ExploitingSA}
Tomas Mikolov, Quoc~V. Le, and Ilya Sutskever.
\newblock Exploiting similarities among languages for machine translation.
\newblock \emph{ArXiv}, abs/1309.4168, 2013{\natexlab{a}}.
\newblock \doi{10.48550/arXiv.1309.4168}.

\bibitem[Fung(1998)]{Fung1998ASV}
Pascale Fung.
\newblock A statistical view on bilingual lexicon extraction.
\newblock In \emph{Proceedings of the Third Conference of the Association for Machine Translation in the Americas: Technical Papers}, pages 1--17. Springer, 1998.

\bibitem[S{\o}gaard et~al.(2018)S{\o}gaard, Ruder, and Vuli{\'c}]{sogaard-etal-2018-limitations}
Anders S{\o}gaard, Sebastian Ruder, and Ivan Vuli{\'c}.
\newblock On the limitations of unsupervised bilingual dictionary induction.
\newblock In \emph{Proceedings of the 56th Annual Meeting of the Association for Computational Linguistics (Volume 1: Long Papers)}, pages 778--788. Association for Computational Linguistics, 2018.
\newblock \doi{10.18653/v1/P18-1072}.

\bibitem[Kovář et~al.(2016)Kovář, Baisa, and Jakubíček]{skengforbillex}
Vojtěch Kovář, Vít Baisa, and Miloš Jakubíček.
\newblock Sketch {E}ngine for bilingual lexicography.
\newblock \emph{International Journal of Lexicography}, 29\penalty0 (3):\penalty0 339–352, 2016.
\newblock \doi{10.1093/ijl/ecw029}.

\bibitem[Baisa et~al.(2016)Baisa, Michelfeit, Medve{\v{d}}, and Jakub{\'\i}{\v{c}}ek]{baisa-etal-2016-european}
V{\'\i}t Baisa, Jan Michelfeit, Marek Medve{\v{d}}, and Milo{\v{s}} Jakub{\'\i}{\v{c}}ek.
\newblock {E}uropean {U}nion language resources in {S}ketch {E}ngine.
\newblock In \emph{Proceedings of the Tenth International Conference on Language Resources and Evaluation ({LREC}'16)}, pages 2799--2803. European Language Resources Association (ELRA), 2016.

\bibitem[Tanaka and Umemura(1994)]{tanaka-umemura-1994-construction}
Kumiko Tanaka and Kyoji Umemura.
\newblock Construction of a bilingual dictionary intermediated by a third language.
\newblock In \emph{{COLING} 1994 Volume 1: The 15th {I}nternational {C}onference on {C}omputational {L}inguistics}, 1994.

\bibitem[Shirai and Yamamoto(2001)]{satoshi}
Satoshi Shirai and Kazuhide Yamamoto.
\newblock Linking english words in two bilingual dictionaries to generate another language pair dictionary.
\newblock 2001.

\bibitem[Saralegi et~al.(2012)Saralegi, Manterola, and San~Vicente]{saralegi-etal-2012-building}
Xabier Saralegi, Iker Manterola, and I{\~n}aki San~Vicente.
\newblock Building a {B}asque-{C}hinese dictionary by using {E}nglish as pivot.
\newblock In \emph{Proceedings of the Eighth International Conference on Language Resources and Evaluation ({LREC}'12)}, pages 1443--1447. European Language Resources Association (ELRA), 2012.

\bibitem[Ordan et~al.(2017)Ordan, Gr{\`a}cia, and Kernerman]{Ordan2017AutogeneratingBD}
Noam Ordan, Jordi Gr{\`a}cia, and Ilan Kernerman.
\newblock Auto-generating bilingual dictionaries.
\newblock In \emph{Proceedings of eLex 2017 conference}, pages 474--484. Electronic lexicography in the 21st century, 2017.

\bibitem[Denisová(2021)]{denisova}
Michaela Denisová.
\newblock Compiling an {E}stonian-{S}lovak dictionary with {E}nglish as a binder.
\newblock In \emph{Proceedings of the eLex 2021 conference}, pages 107--120. Lexical Computing CZ, s.r.o., 2021.

\bibitem[Denisová(2022)]{denisova-2022}
Michaela Denisová.
\newblock Parallel, or comparable? {T}hat is the question: {T}he comparison of parallel and comparable data-based methods for bilingual lexicon induction.
\newblock In \emph{Proceedings of the Sixteenth Workshop on Recent Advances in Slavonic Natural Languages Processing (RASLAN)}, pages 4--13. Tribun EU, 2022.

\bibitem[Felix(2020)]{Stahlberg2020NeuralMT}
Stahlberg Felix.
\newblock Neural machine translation: A review.
\newblock \emph{The Journal of Artificial Intelligence Research}, 69:\penalty0 343--418, 2020.
\newblock \doi{10.1613/jair.1.12007}.

\bibitem[Klein et~al.(2017)Klein, Kim, Deng, Senellart, and Rush]{klein-etal-2017-opennmt}
Guillaume Klein, Yoon Kim, Yuntian Deng, Jean Senellart, and Alexander Rush.
\newblock {O}pen{NMT}: Open-source toolkit for neural machine translation.
\newblock In \emph{Proceedings of {ACL} 2017, System Demonstrations}, pages 67--72. Association for Computational Linguistics, 2017.
\newblock URL \url{https://aclanthology.org/P17-4012}.

\bibitem[Vaswani et~al.(2018)Vaswani, Bengio, Brevdo, Chollet, Gomez, Gouws, Jones, Kaiser, Kalchbrenner, Parmar, Sepassi, Shazeer, and Uszkoreit]{tensor2tensor}
Ashish Vaswani, Samy Bengio, Eugene Brevdo, Francois Chollet, Aidan Gomez, Stephan Gouws, Llion Jones, {\L}ukasz Kaiser, Nal Kalchbrenner, Niki Parmar, Ryan Sepassi, Noam Shazeer, and Jakob Uszkoreit.
\newblock {T}ensor2{T}ensor for neural machine translation.
\newblock In \emph{Proceedings of the 13th Conference of the Association for Machine Translation in the {A}mericas (Volume 1: Research Track)}, pages 193--199. Association for Machine Translation in the Americas, 2018.

\bibitem[Ott et~al.(2019)Ott, Edunov, Baevski, Fan, Gross, Ng, Grangier, and Auli]{ott2019fairseq}
Myle Ott, Sergey Edunov, Alexei Baevski, Angela Fan, Sam Gross, Nathan Ng, David Grangier, and Michael Auli.
\newblock fairseq: A fast, extensible toolkit for sequence modeling.
\newblock In \emph{Proceedings of the 2019 Conference of the North {A}merican Chapter of the Association for Computational Linguistics (Demonstrations)}, pages 48--53. Association for Computational Linguistics, 2019.
\newblock \doi{10.18653/v1/N19-4009}.

\bibitem[Lewis et~al.(2020)Lewis, Liu, Goyal, Ghazvininejad, Mohamed, Levy, Stoyanov, and Zettlemoyer]{lewis-etal-2020-bart}
Mike Lewis, Yinhan Liu, Naman Goyal, Marjan Ghazvininejad, Abdelrahman Mohamed, Omer Levy, Veselin Stoyanov, and Luke Zettlemoyer.
\newblock {BART}: Denoising sequence-to-sequence pre-training for natural language generation, translation, and comprehension.
\newblock In \emph{Proceedings of the 58th Annual Meeting of the Association for Computational Linguistics}, pages 7871--7880. Association for Computational Linguistics, 2020.
\newblock \doi{10.18653/v1/2020.acl-main.703}.

\bibitem[Dinu and Baroni(2014)]{Dinu2015ImprovingZL}
Georgiana Dinu and Marco Baroni.
\newblock Improving zero-shot learning by mitigating the hubness problem.
\newblock \emph{CoRR}, abs/1412.6568, 2014.
\newblock \doi{/10.48550/arXiv.1412.6568}.

\bibitem[Glava{\v{s}} et~al.(2019)Glava{\v{s}}, Litschko, Ruder, and Vuli{\'c}]{glavas-etal-2019-properly}
Goran Glava{\v{s}}, Robert Litschko, Sebastian Ruder, and Ivan Vuli{\'c}.
\newblock How to (properly) evaluate cross-lingual word embeddings: On strong baselines, comparative analyses, and some misconceptions.
\newblock In \emph{Proceedings of the 57th Annual Meeting of the Association for Computational Linguistics}, pages 710--721. Association for Computational Linguistics, 2019.
\newblock \doi{10.18653/v1/P19-1070}.

\bibitem[Vuli{\'c} et~al.(2019)Vuli{\'c}, Glava{\v{s}}, Reichart, and Korhonen]{vulic-etal-2019-really}
Ivan Vuli{\'c}, Goran Glava{\v{s}}, Roi Reichart, and Anna Korhonen.
\newblock Do we really need fully unsupervised cross-lingual embeddings?
\newblock In \emph{Proceedings of the 2019 Conference on Empirical Methods in Natural Language Processing and the 9th International Joint Conference on Natural Language Processing (EMNLP-IJCNLP)}, pages 4407--4418. Association for Computational Linguistics, 2019.
\newblock \doi{10.18653/v1/D19-1449}.

\bibitem[Wang et~al.(2022)Wang, Wu, He, Huang, and Church]{Wang2021ProgressIM}
Haifeng Wang, Hua Wu, Zhongjun He, Liang Huang, and Kenneth~Ward Church.
\newblock Progress in machine translation.
\newblock \emph{Engineering}, 18:\penalty0 143--153, 2022.
\newblock \doi{10.1016/j.eng.2021.03.023}.

\bibitem[Vaswani et~al.(2017)Vaswani, Shazeer, Parmar, Uszkoreit, Jones, Gomez, Kaiser, and Polosukhin]{Vaswani2017AttentionIA}
Ashish Vaswani, Noam~M. Shazeer, Niki Parmar, Jakob Uszkoreit, Llion Jones, Aidan~N. Gomez, Lukasz Kaiser, and Illia Polosukhin.
\newblock Attention is all you need.
\newblock In \emph{Neural Information Processing Systems}, 2017.
\newblock \doi{10.48550/arXiv.1706.03762}.

\bibitem[Devlin et~al.(2019)Devlin, Chang, Lee, and Toutanova]{devlin-etal-2019-bert}
Jacob Devlin, Ming-Wei Chang, Kenton Lee, and Kristina Toutanova.
\newblock {BERT}: Pre-training of deep bidirectional transformers for language understanding.
\newblock In \emph{Proceedings of the 2019 Conference of the North {A}merican Chapter of the Association for Computational Linguistics: Human Language Technologies, Volume 1 (Long and Short Papers)}, pages 4171--4186. Association for Computational Linguistics, 2019.
\newblock \doi{10.18653/v1/N19-1423}.

\bibitem[Raffel et~al.(2020)Raffel, Shazeer, Roberts, Lee, Narang, Matena, Zhou, Li, and Liu]{DBLP:journals/corr/abs-1910-10683}
Colin Raffel, Noam~M. Shazeer, Adam Roberts, Katherine Lee, Sharan Narang, Michael Matena, Yanqi Zhou, Wei Li, and Peter~J. Liu.
\newblock Exploring the limits of transfer learning with a unified text-to-text transformer.
\newblock \emph{Journal of Machine Learning Research}, 21:\penalty0 1--67, 2020.

\bibitem[Tiedemann and Thottingal(2020)]{tiedemann-thottingal-2020-opus}
J{\"o}rg Tiedemann and Santhosh Thottingal.
\newblock {OPUS}-{MT} {--} building open translation services for the world.
\newblock In \emph{Proceedings of the 22nd Annual Conference of the European Association for Machine Translation}, pages 479--480. European Association for Machine Translation, 2020.

\bibitem[Liu et~al.(2020)Liu, Gu, Goyal, Li, Edunov, Ghazvininejad, Lewis, and Zettlemoyer]{liu-etal-2020-multilingual-denoising}
Yinhan Liu, Jiatao Gu, Naman Goyal, Xian Li, Sergey Edunov, Marjan Ghazvininejad, Mike Lewis, and Luke Zettlemoyer.
\newblock Multilingual denoising pre-training for neural machine translation.
\newblock \emph{Transactions of the Association for Computational Linguistics}, 8:\penalty0 726--742, 2020.
\newblock \doi{10.1162/tacl_a_00343}.

\bibitem[Tiedemann(2012)]{tiedemann-2012-parallel}
J{\"o}rg Tiedemann.
\newblock Parallel data, tools and interfaces in {OPUS}.
\newblock In \emph{Proceedings of the Eighth International Conference on Language Resources and Evaluation ({LREC}'12)}, pages 2214--2218. European Language Resources Association (ELRA), 2012.

\bibitem[Xue et~al.(2021)Xue, Constant, Roberts, Kale, Al-Rfou, Siddhant, Barua, and Raffel]{xue-etal-2021-mt5}
Linting Xue, Noah Constant, Adam Roberts, Mihir Kale, Rami Al-Rfou, Aditya Siddhant, Aditya Barua, and Colin Raffel.
\newblock m{T}5: A massively multilingual pre-trained text-to-text transformer.
\newblock In \emph{Proceedings of the 2021 Conference of the North American Chapter of the Association for Computational Linguistics: Human Language Technologies}, pages 483--498. Association for Computational Linguistics, 2021.
\newblock \doi{10.18653/v1/2021.naacl-main.41}.

\bibitem[Rapp(1995)]{rapp95}
Reinhard Rapp.
\newblock Identifying word translations in non-parallel texts.
\newblock In \emph{33rd Annual Meeting of the Association for Computational Linguistics}, pages 320--322. Association for Computational Linguistics, 1995.
\newblock \doi{10.3115/981658.981709}.

\bibitem[Rapp(1999)]{rapp99}
Reinhard Rapp.
\newblock Automatic identification of word translations from unrelated {E}nglish and {G}erman corpora.
\newblock In \emph{Proceedings of the 37th Annual Meeting of the Association for Computational Linguistics}, pages 519--526. Association for Computational Linguistics, 1999.
\newblock \doi{10.3115/1034678.1034756}.

\bibitem[Haghighi et~al.(2008)Haghighi, Liang, Berg-Kirkpatrick, and Klein]{Haghighi2008LearningBL}
Aria Haghighi, Percy Liang, Taylor Berg-Kirkpatrick, and Dan Klein.
\newblock Learning bilingual lexicons from monolingual corpora.
\newblock In \emph{Proceedings of ACL-08: HLT}, pages 771--779. Association for Computational Linguistics, 2008.

\bibitem[Schafer and Yarowsky(2002)]{schafyaro02}
Charles Schafer and David Yarowsky.
\newblock Inducing translation lexicons via diverse similarity measures and bridge languages.
\newblock In \emph{{COLING}-02: The 6th Conference on Natural Language Learning 2002 ({C}o{NLL}-2002)}, 2002.

\bibitem[Koehn and Knight(2002)]{kk02}
Philipp Koehn and Kevin Knight.
\newblock Learning a translation lexicon from monolingual corpora.
\newblock In \emph{Proceedings of the {ACL}-02 Workshop on Unsupervised Lexical Acquisition}, pages 9--16. Association for Computational Linguistics, 2002.
\newblock \doi{10.3115/1118627.1118629}.
\newblock URL \url{https://aclanthology.org/W02-0902}.

\bibitem[Gaussier et~al.(2004)Gaussier, Renders, Matveeva, Goutte, and Dejean]{Gaussier2004AGV}
Eric Gaussier, J.M. Renders, I.~Matveeva, C.~Goutte, and H.~Dejean.
\newblock A geometric view on bilingual lexicon extraction from comparable corpora.
\newblock In \emph{Proceedings of the 42nd Annual Meeting of the Association for Computational Linguistics ({ACL}-04)}, pages 526--533, 2004.
\newblock \doi{10.3115/1218955.1219022}.

\bibitem[Artetxe et~al.(2018{\natexlab{a}})Artetxe, Labaka, and Agirre]{artetxe-etal-2018-robust}
Mikel Artetxe, Gorka Labaka, and Eneko Agirre.
\newblock A robust self-learning method for fully unsupervised cross-lingual mappings of word embeddings.
\newblock In \emph{Proceedings of the 56th Annual Meeting of the Association for Computational Linguistics (Volume 1: Long Papers)}, pages 789--798. Association for Computational Linguistics, 2018{\natexlab{a}}.
\newblock \doi{10.18653/v1/P18-1073}.

\bibitem[Kementchedjhieva et~al.(2018)Kementchedjhieva, Ruder, Cotterell, and S{\o}gaard]{kementchedjhieva-etal-2020-generalizing}
Yova Kementchedjhieva, Sebastian Ruder, Ryan Cotterell, and Anders S{\o}gaard.
\newblock Generalizing {P}rocrustes analysis for better bilingual dictionary induction.
\newblock In \emph{Proceedings of the 22nd Conference on Computational Natural Language Learning}, pages 211--220. Association for Computational Linguistics, 2018.
\newblock \doi{10.18653/v1/K18-1021}.

\bibitem[Woller et~al.(2021)Woller, Hangya, and Fraser]{woller-etal-2021-neglect}
Lisa Woller, Viktor Hangya, and Alexander Fraser.
\newblock Do not neglect related languages: The case of low-resource {O}ccitan cross-lingual word embeddings.
\newblock In \emph{Proceedings of the 1st Workshop on Multilingual Representation Learning}, pages 41--50. Association for Computational Linguistics, 2021.
\newblock \doi{10.18653/v1/2021.mrl-1.4}.

\bibitem[Bai et~al.(2019)Bai, Zhang, Cao, and Zhao]{Bai2022DualityRF}
Xuefeng Bai, Yue Zhang, Hailong Cao, and Tiejun Zhao.
\newblock Duality regularization for unsupervised bilingual lexicon induction.
\newblock \emph{ArXiv}, abs/1909.01013, 2019.

\bibitem[Marchisio et~al.(2022)Marchisio, Saad-Eldin, Duh, Priebe, and Koehn]{Marchisio2022BilingualLI}
Kelly Marchisio, Ali Saad-Eldin, Kevin Duh, Carey Priebe, and Philipp Koehn.
\newblock Bilingual lexicon induction for low-resource languages using graph matching via optimal transport.
\newblock In \emph{Proceedings of the 2022 Conference on Empirical Methods in Natural Language Processing}, pages 2545--2561. Association for Computational Linguistics, 2022.
\newblock \doi{10.18653/v1/2022.emnlp-main.164}.

\bibitem[Ruder et~al.(2019{\natexlab{a}})Ruder, Vuli{\'c}, and S{\o}gaard]{Ruder2017ASO}
Sebastian Ruder, Ivan Vuli{\'c}, and Anders S{\o}gaard.
\newblock A survey of cross-lingual word embedding models.
\newblock \emph{The Journal of Artificial Intelligence Research}, 65:\penalty0 569--631, 2019{\natexlab{a}}.
\newblock \doi{10.1613/jair.1.11640}.

\bibitem[Smith(2019)]{Smith2019ContextualWR}
Noah~A. Smith.
\newblock Contextual word representations: A contextual introduction.
\newblock \emph{ArXiv}, abs/1902.06006, 2019.
\newblock \doi{10.48550/arXiv.1902.06006}.

\bibitem[Bojanowski et~al.(2017)Bojanowski, Grave, Joulin, and Mikolov]{bojanowski-etal-2017-enriching}
Piotr Bojanowski, Edouard Grave, Armand Joulin, and Tomas Mikolov.
\newblock Enriching word vectors with subword information.
\newblock \emph{Transactions of the Association for Computational Linguistics}, 5:\penalty0 135--146, 2017.
\newblock \doi{10.1162/tacl_a_00051}.

\bibitem[Ren et~al.(2020)Ren, Liu, Zhou, and Ma]{ren-etal-2020-graph}
Shuo Ren, Shujie Liu, Ming Zhou, and Shuai Ma.
\newblock A graph-based coarse-to-fine method for unsupervised bilingual lexicon induction.
\newblock In \emph{Proceedings of the 58th Annual Meeting of the Association for Computational Linguistics}, pages 3476--3485. Association for Computational Linguistics, 2020.
\newblock \doi{10.18653/v1/2020.acl-main.318}.

\bibitem[Riley and Gildea(2020)]{Riley2020UnsupervisedBL}
Parker Riley and Daniel Gildea.
\newblock Unsupervised bilingual lexicon induction across writing systems.
\newblock \emph{ArXiv}, abs/2002.00037, 2020.
\newblock \doi{10.48550/arXiv.2002.00037}.

\bibitem[Mohiuddin et~al.(2020)Mohiuddin, Bari, and Joty]{mohiuddin-etal-2020-lnmap}
Tasnim Mohiuddin, M~Saiful Bari, and Shafiq Joty.
\newblock {LNM}ap: Departures from isomorphic assumption in bilingual lexicon induction through non-linear mapping in latent space.
\newblock In \emph{Proceedings of the 2020 Conference on Empirical Methods in Natural Language Processing (EMNLP)}, pages 2712--2723. Association for Computational Linguistics, 2020.
\newblock \doi{10.18653/v1/2020.emnlp-main.215}.

\bibitem[Vuli{\'c} et~al.(2020)Vuli{\'c}, Korhonen, and Glava{\v{s}}]{vulic-etal-2020-improving}
Ivan Vuli{\'c}, Anna Korhonen, and Goran Glava{\v{s}}.
\newblock Improving bilingual lexicon induction with unsupervised post-processing of monolingual word vector spaces.
\newblock In \emph{Proceedings of the 5th Workshop on Representation Learning for NLP}, pages 45--54. Association for Computational Linguistics, 2020.
\newblock \doi{10.18653/v1/2020.repl4nlp-1.7}.

\bibitem[Glava{\v{s}} and Vuli{\'c}(2020)]{glavas-vulic-2020-non}
Goran Glava{\v{s}} and Ivan Vuli{\'c}.
\newblock Non-linear instance-based cross-lingual mapping for non-isomorphic embedding spaces.
\newblock In \emph{Proceedings of the 58th Annual Meeting of the Association for Computational Linguistics}, pages 7548--7555. Association for Computational Linguistics, 2020.
\newblock \doi{10.18653/v1/2020.acl-main.675}.

\bibitem[Mikolov et~al.(2013{\natexlab{b}})Mikolov, Chen, Corrado, and Dean]{Mikolov2013EfficientEO}
Tomas Mikolov, Kai Chen, Gregory~S. Corrado, and Jeffrey Dean.
\newblock Efficient estimation of word representations in vector space.
\newblock In \emph{International Conference on Learning Representations}, 2013{\natexlab{b}}.
\newblock \doi{10.48550/arXiv.1301.3781}.

\bibitem[Severini et~al.(2020)Severini, Hangya, Fraser, and Sch{\"u}tze]{severini-etal-2020-lmu}
Silvia Severini, Viktor Hangya, Alexander Fraser, and Hinrich Sch{\"u}tze.
\newblock {LMU} bilingual dictionary induction system with word surface similarity scores for {BUCC} 2020.
\newblock In \emph{Proceedings of the 13th Workshop on Building and Using Comparable Corpora}, pages 49--55. European Language Resources Association, 2020.

\bibitem[Michel et~al.(2020)Michel, Hangya, and Fraser]{michel-etal-2020-exploring}
Leah Michel, Viktor Hangya, and Alexander Fraser.
\newblock Exploring bilingual word embeddings for {H}iligaynon, a low-resource language.
\newblock In \emph{Proceedings of the Twelfth Language Resources and Evaluation Conference}, pages 2573--2580. European Language Resources Association, 2020.

\bibitem[Laville et~al.(2020)Laville, Hazem, and Morin]{laville-etal-2020-taln}
Martin Laville, Amir Hazem, and Emmanuel Morin.
\newblock {TALN}/{LS}2{N} participation at the {BUCC} shared task: Bilingual dictionary induction from comparable corpora.
\newblock In \emph{Proceedings of the 13th Workshop on Building and Using Comparable Corpora}, pages 56--60. European Language Resources Association, 2020.

\bibitem[Eder et~al.(2021)Eder, Hangya, and Fraser]{eder-etal-2021-anchor}
Tobias Eder, Viktor Hangya, and Alexander Fraser.
\newblock Anchor-based bilingual word embeddings for low-resource languages.
\newblock In \emph{Proceedings of the 59th Annual Meeting of the Association for Computational Linguistics and the 11th International Joint Conference on Natural Language Processing (Volume 2: Short Papers)}, pages 227--232. Association for Computational Linguistics, 2021.
\newblock \doi{10.18653/v1/2021.acl-short.30}.

\bibitem[Chakravarthi et~al.(2020)Chakravarthi, Rajasekaran, Arcan, McGuinness, E.~O{'}Connor, and McCrae]{chakravarthi-etal-2020-bilingual}
Bharathi~Raja Chakravarthi, Navaneethan Rajasekaran, Mihael Arcan, Kevin McGuinness, Noel E.~O{'}Connor, and John~P. McCrae.
\newblock Bilingual lexicon induction across orthographically-distinct under-resourced {D}ravidian languages.
\newblock In \emph{Proceedings of the 7th Workshop on NLP for Similar Languages, Varieties and Dialects}, pages 57--69. International Committee on Computational Linguistics (ICCL), 2020.

\bibitem[Marchisio et~al.(2021{\natexlab{a}})Marchisio, Park, Saad-Eldin, Alyakin, Duh, Priebe, and Koehn]{marchisio-etal-2021-analysis-euclidean}
Kelly Marchisio, Youngser Park, Ali Saad-Eldin, Anton Alyakin, Kevin Duh, Carey Priebe, and Philipp Koehn.
\newblock An analysis of {E}uclidean vs. graph-based framing for bilingual lexicon induction from word embedding spaces.
\newblock In \emph{Findings of the Association for Computational Linguistics: EMNLP 2021}, pages 738--749. Association for Computational Linguistics, 2021{\natexlab{a}}.
\newblock \doi{10.18653/v1/2021.findings-emnlp.64}.

\bibitem[Sannigrahi and Read(2022)]{Sannigrahi2022IsomorphicCE}
Sonal Sannigrahi and Jesse Read.
\newblock Isomorphic cross-lingual embeddings for low-resource languages.
\newblock \emph{ArXiv}, abs/2203.14632, 2022.
\newblock \doi{10.48550/arXiv.2203.14632}.

\bibitem[Baroni et~al.(2009)Baroni, Bernardini, Ferraresi, and Zanchetta]{Baroni2009TheWW}
Marco Baroni, Silvia Bernardini, Adriano Ferraresi, and Eros Zanchetta.
\newblock The wacky wide web: a collection of very large linguistically processed web-crawled corpora.
\newblock \emph{Language Resources and Evaluation}, 43:\penalty0 209--226, 2009.
\newblock \doi{/10.1007/s10579-009-9081-4}.

\bibitem[Edmiston et~al.(2022)Edmiston, Keung, and Smith]{edmiston-etal-2022-domain}
Daniel Edmiston, Phillip Keung, and Noah~A. Smith.
\newblock Domain mismatch doesn{'}t always prevent cross-lingual transfer learning.
\newblock In \emph{Proceedings of the Thirteenth Language Resources and Evaluation Conference}, pages 892--899. European Language Resources Association, 2022.

\bibitem[Dita et~al.(2009)Dita, Roxas, and Inventado]{dita-etal-2009-building}
Shirley~N. Dita, Rachel Edita~O. Roxas, and Paul Inventado.
\newblock Building online corpora of {P}hilippine languages.
\newblock In \emph{Proceedings of the 23rd Pacific Asia Conference on Language, Information and Computation, Volume 2}, pages 646--653. City University of Hong Kong, 2009.

\bibitem[Vuli{\'c} and Korhonen(2016)]{vulic-korhonen-2016-role}
Ivan Vuli{\'c} and Anna Korhonen.
\newblock On the role of seed lexicons in learning bilingual word embeddings.
\newblock In \emph{Proceedings of the 54th Annual Meeting of the Association for Computational Linguistics (Volume 1: Long Papers)}, pages 247--257. Association for Computational Linguistics, 2016.
\newblock \doi{10.18653/v1/P16-1024}.

\bibitem[Izbicki(2022)]{izbicki-2022-aligning}
Mike Izbicki.
\newblock Aligning word vectors on low-resource languages with {W}iktionary.
\newblock In \emph{Proceedings of the Fifth Workshop on Technologies for Machine Translation of Low-Resource Languages (LoResMT 2022)}, pages 107--117. Association for Computational Linguistics, 2022.

\bibitem[Adjali et~al.(2022)Adjali, Morin, Sharoff, Rapp, and Zweigenbaum]{Adjali2022OverviewOT}
Omar Adjali, Emmanuel Morin, Serge Sharoff, Reinhard Rapp, and Pierre Zweigenbaum.
\newblock {Overview of the 2022 BUCC Shared Task: Bilingual Term Alignment in Comparable Specialized Corpora}.
\newblock In \emph{{BUCC, 15th Workshop on Building and Using Comparable Corpora}}, pages 67--76, 2022.

\bibitem[Wang et~al.(2020)Wang, Hou, Che, and Liu]{Wang2020FromST}
Yuxuan Wang, Yutai Hou, Wanxiang Che, and Ting Liu.
\newblock From static to dynamic word representations: a survey.
\newblock \emph{International Journal of Machine Learning and Cybernetics}, 11:\penalty0 1611--1630, 2020.
\newblock \doi{10.1007/s13042-020-01069-8}.

\bibitem[Lample et~al.(2017)Lample, Denoyer, and Ranzato]{Lample2017UnsupervisedMT}
Guillaume Lample, Ludovic Denoyer, and Marc'Aurelio Ranzato.
\newblock Unsupervised machine translation using monolingual corpora only.
\newblock \emph{ArXiv}, abs/1711.00043, 2017.
\newblock \doi{/10.48550/arXiv.1711.00043}.

\bibitem[Artetxe et~al.(2016)Artetxe, Labaka, and Agirre]{artetxe-etal-2016-learning}
Mikel Artetxe, Gorka Labaka, and Eneko Agirre.
\newblock Learning principled bilingual mappings of word embeddings while preserving monolingual invariance.
\newblock In \emph{Proceedings of the 2016 Conference on Empirical Methods in Natural Language Processing}, pages 2289--2294. Association for Computational Linguistics, 2016.
\newblock \doi{10.18653/v1/D16-1250}.

\bibitem[Artetxe et~al.(2017{\natexlab{b}})Artetxe, Labaka, and Agirre]{artetxe-etal-2017-learning}
Mikel Artetxe, Gorka Labaka, and Eneko Agirre.
\newblock Learning bilingual word embeddings with (almost) no bilingual data.
\newblock In \emph{Proceedings of the 55th Annual Meeting of the Association for Computational Linguistics (Volume 1: Long Papers)}, pages 451--462. Association for Computational Linguistics, 2017{\natexlab{b}}.
\newblock \doi{10.18653/v1/P17-1042}.

\bibitem[Artetxe et~al.(2018{\natexlab{b}})Artetxe, Labaka, and Agirre]{artetxe2018aaai}
Mikel Artetxe, Gorka Labaka, and Eneko Agirre.
\newblock Generalizing and improving bilingual word embedding mappings with a multi-step framework of linear transformations.
\newblock \emph{Proceedings of the AAAI Conference on Artificial Intelligence}, 32\penalty0 (1), 2018{\natexlab{b}}.
\newblock \doi{10.1609/aaai.v32i1.11992}.

\bibitem[Joulin et~al.(2018)Joulin, Bojanowski, Mikolov, J{\'e}gou, and Grave]{joulin-etal-2018-loss}
Armand Joulin, Piotr Bojanowski, Tomas Mikolov, Herv{\'e} J{\'e}gou, and Edouard Grave.
\newblock Loss in translation: Learning bilingual word mapping with a retrieval criterion.
\newblock In \emph{Proceedings of the 2018 Conference on Empirical Methods in Natural Language Processing}, pages 2979--2984. Association for Computational Linguistics, 2018.
\newblock \doi{10.18653/v1/D18-1330}.

\bibitem[Radovanović et~al.(2010)Radovanović, Nanopoulos, and Ivanovi{\'c}]{Radovanovi2010TimeSeriesCI}
Miloš Radovanović, Alexandros Nanopoulos, and Mirjana Ivanovi{\'c}.
\newblock Hubs in space: Popular nearest neighbors in high-dimensional data.
\newblock \emph{The Journal of Machine Learning Research}, 11:\penalty0 2487--2531, 2010.
\newblock \doi{10.5555/1756006.1953015}.

\bibitem[Kementchedjhieva et~al.(2019)Kementchedjhieva, Hartmann, and S{\o}gaard]{kementchedjhieva-etal-2019-lost}
Yova Kementchedjhieva, Mareike Hartmann, and Anders S{\o}gaard.
\newblock Lost in evaluation: Misleading benchmarks for bilingual dictionary induction.
\newblock In \emph{Proceedings of the 2019 Conference on Empirical Methods in Natural Language Processing and the 9th International Joint Conference on Natural Language Processing (EMNLP-IJCNLP)}, pages 3336--3341. Association for Computational Linguistics, 2019.
\newblock \doi{10.18653/v1/D19-1328}.

\bibitem[Nishikawa et~al.(2021)Nishikawa, Ri, and Tsuruoka]{nishikawa-etal-2021-data}
Sosuke Nishikawa, Ryokan Ri, and Yoshimasa Tsuruoka.
\newblock Data augmentation with unsupervised machine translation improves the structural similarity of cross-lingual word embeddings.
\newblock In \emph{Proceedings of the 59th Annual Meeting of the Association for Computational Linguistics and the 11th International Joint Conference on Natural Language Processing: Student Research Workshop}, pages 163--173. Association for Computational Linguistics, 2021.
\newblock \doi{10.18653/v1/2021.acl-srw.17}.

\bibitem[Ruder et~al.(2019{\natexlab{b}})Ruder, S{\o}gaard, and Vuli{\'c}]{ruder-etal-2019-unsupervised}
Sebastian Ruder, Anders S{\o}gaard, and Ivan Vuli{\'c}.
\newblock Unsupervised cross-lingual representation learning.
\newblock In \emph{Proceedings of the 57th Annual Meeting of the Association for Computational Linguistics: Tutorial Abstracts}, pages 31--38. Association for Computational Linguistics, 2019{\natexlab{b}}.
\newblock \doi{10.18653/v1/P19-4007}.

\bibitem[Kondrak et~al.(2003)Kondrak, Marcu, and Knight]{kondrak-etal-2003-cognates}
Grzegorz Kondrak, Daniel Marcu, and Kevin Knight.
\newblock Cognates can improve statistical translation models.
\newblock In \emph{Companion Volume of the Proceedings of {HLT}-{NAACL} 2003 - Short Papers}, pages 46--48, 2003.

\bibitem[Artetxe et~al.(2018{\natexlab{c}})Artetxe, Labaka, Lopez-Gazpio, and Agirre]{artetxe-etal-2018-uncovering}
Mikel Artetxe, Gorka Labaka, I{\~n}igo Lopez-Gazpio, and Eneko Agirre.
\newblock Uncovering divergent linguistic information in word embeddings with lessons for intrinsic and extrinsic evaluation.
\newblock In \emph{Proceedings of the 22nd Conference on Computational Natural Language Learning}, pages 282--291. Association for Computational Linguistics, 2018{\natexlab{c}}.
\newblock \doi{10.18653/v1/K18-1028}.

\bibitem[Riley and Gildea(2018)]{riley-gildea-2018-orthographic}
Parker Riley and Daniel Gildea.
\newblock Orthographic features for bilingual lexicon induction.
\newblock In \emph{Proceedings of the 56th Annual Meeting of the Association for Computational Linguistics (Volume 2: Short Papers)}, pages 390--394. Association for Computational Linguistics, 2018.
\newblock \doi{10.18653/v1/P18-2062}.

\bibitem[Hulden et~al.(2014)Hulden, Forsberg, and Ahlberg]{hulden-etal-2014-semi}
Mans Hulden, Markus Forsberg, and Malin Ahlberg.
\newblock Semi-supervised learning of morphological paradigms and lexicons.
\newblock In \emph{Proceedings of the 14th Conference of the {E}uropean Chapter of the Association for Computational Linguistics}, pages 569--578. Association for Computational Linguistics, 2014.
\newblock \doi{10.3115/v1/E14-1060}.

\bibitem[Sorokin(2016)]{sorokin-2016-using}
Alexey Sorokin.
\newblock Using longest common subsequence and character models to predict word forms.
\newblock In \emph{Proceedings of the 14th {SIGMORPHON} Workshop on Computational Research in Phonetics, Phonology, and Morphology}, pages 54--61. Association for Computational Linguistics, 2016.
\newblock \doi{10.18653/v1/W16-2009}.

\bibitem[Bhattacharyya(2010)]{pushpak_2010}
Pushpak Bhattacharyya.
\newblock {I}ndo{W}ord{N}et.
\newblock In \emph{Proceedings of the Seventh International Conference on Language Resources and Evaluation ({LREC}'10)}. European Language Resources Association (ELRA), 2010.

\bibitem[Braune et~al.(2018)Braune, Hangya, Eder, and Fraser]{braune-etal-2018-evaluating}
Fabienne Braune, Viktor Hangya, Tobias Eder, and Alexander Fraser.
\newblock Evaluating bilingual word embeddings on the long tail.
\newblock In \emph{Proceedings of the 2018 Conference of the North {A}merican Chapter of the Association for Computational Linguistics: Human Language Technologies, Volume 2 (Short Papers)}, pages 188--193. Association for Computational Linguistics, 2018.
\newblock \doi{10.18653/v1/N18-2030}.

\bibitem[Goodfellow et~al.(2014)Goodfellow, Pouget-Abadie, Mirza, Xu, Warde-Farley, Ozair, Courville, and Bengio]{Goodfellow2014GenerativeAN}
Ian Goodfellow, Jean Pouget-Abadie, Mehdi Mirza, Bing Xu, David Warde-Farley, Sherjil Ozair, Aaron Courville, and Yoshua Bengio.
\newblock Generative adversarial nets.
\newblock In \emph{Advances in Neural Information Processing Systems}, volume~27. Neural Information Processing Systems, 2014.

\bibitem[Li et~al.(2021)Li, Zhang, Yu, and Hu]{Li2021AdversarialTW}
Yuling Li, Yuhong Zhang, Kui Yu, and Xuegang Hu.
\newblock Adversarial training with wasserstein distance for learning cross-lingual word embeddings.
\newblock \emph{Appl Intell}, 51:\penalty0 7666--7678, 2021.
\newblock \doi{/10.1007/s10489-020-02136-x}.

\bibitem[Zhao et~al.(2020)Zhao, Wang, Zhang, and Wu]{zhao-etal-2020-relaxed}
Xu~Zhao, Zihao Wang, Yong Zhang, and Hao Wu.
\newblock A relaxed matching procedure for unsupervised {BLI}.
\newblock In \emph{Proceedings of the 58th Annual Meeting of the Association for Computational Linguistics}, pages 3036--3041. Association for Computational Linguistics, 2020.
\newblock \doi{10.18653/v1/2020.acl-main.274}.

\bibitem[Grave et~al.(2018)Grave, Joulin, and Berthet]{Grave2018UnsupervisedAO}
Edouard Grave, Armand Joulin, and Quentin Berthet.
\newblock Unsupervised alignment of embeddings with wasserstein procrustes.
\newblock \emph{ArXiv}, abs/1805.11222, 2018.
\newblock \doi{/10.48550/arXiv.1805.11222}.

\bibitem[Feng et~al.(2022)Feng, Cao, Zhao, Wang, and Peng]{feng-etal-2022-cross}
Zihao Feng, Hailong Cao, Tiejun Zhao, Weixuan Wang, and Wei Peng.
\newblock Cross-lingual feature extraction from monolingual corpora for low-resource unsupervised bilingual lexicon induction.
\newblock In \emph{Proceedings of the 29th International Conference on Computational Linguistics}, pages 5278--5287. International Committee on Computational Linguistics, 2022.

\bibitem[Hartmann et~al.(2019)Hartmann, Kementchedjhieva, and S{\o}gaard]{hartmann2019unsup}
Mareike Hartmann, Yova Kementchedjhieva, and Anders S{\o}gaard.
\newblock Comparing unsupervised word translation methods step by step.
\newblock In \emph{Neural Information Processing Systems}, 2019.

\bibitem[Karan et~al.(2020)Karan, Vuli{\'c}, Korhonen, and Glava{\v{s}}]{karan-etal-2020-classification}
Mladen Karan, Ivan Vuli{\'c}, Anna Korhonen, and Goran Glava{\v{s}}.
\newblock Classification-based self-learning for weakly supervised bilingual lexicon induction.
\newblock In \emph{Proceedings of the 58th Annual Meeting of the Association for Computational Linguistics}, pages 6915--6922. Association for Computational Linguistics, 2020.
\newblock \doi{10.18653/v1/2020.acl-main.618}.

\bibitem[Luong et~al.(2015)Luong, Pham, and Manning]{luong-etal-2015-bilingual}
Thang Luong, Hieu Pham, and Christopher~D. Manning.
\newblock Bilingual word representations with monolingual quality in mind.
\newblock In \emph{Proceedings of the 1st Workshop on Vector Space Modeling for Natural Language Processing}, pages 151--159. Association for Computational Linguistics, 2015.
\newblock \doi{10.3115/v1/W15-1521}.

\bibitem[Pavlick et~al.(2014)Pavlick, Post, Irvine, Kachaev, and Callison-Burch]{pavlick-etal-2014-language}
Ellie Pavlick, Matt Post, Ann Irvine, Dmitry Kachaev, and Chris Callison-Burch.
\newblock The language demographics of {A}mazon {M}echanical {T}urk.
\newblock \emph{Transactions of the Association for Computational Linguistics}, 2:\penalty0 79--92, 2014.
\newblock \doi{10.1162/tacl_a_00167}.

\bibitem[Cao et~al.(2023)Cao, Zhao, Wang, and Peng]{Cao2023BilingualWE}
Hailong Cao, Tiejun Zhao, Weixuan Wang, and Wei Peng.
\newblock Bilingual word embedding fusion for robust unsupervised bilingual lexicon induction.
\newblock \emph{Information Fusion}, 97:\penalty0 101818, 2023.
\newblock \doi{/10.1016/j.inffus.2023.101818}.

\bibitem[Artetxe et~al.(2019{\natexlab{b}})Artetxe, Labaka, and Agirre]{artetxe-etal-2019-bilingual}
Mikel Artetxe, Gorka Labaka, and Eneko Agirre.
\newblock Bilingual lexicon induction through unsupervised machine translation.
\newblock In \emph{Proceedings of the 57th Annual Meeting of the Association for Computational Linguistics}, pages 5002--5007. Association for Computational Linguistics, 2019{\natexlab{b}}.
\newblock \doi{10.18653/v1/P19-1494}.

\bibitem[Chang et~al.(2008)Chang, Ratinov, Roth, and Srikumar]{Chang2008ImportanceOS}
Ming-Wei Chang, Lev-Arie Ratinov, Dan Roth, and Vivek Srikumar.
\newblock Importance of semantic representation: Dataless classification.
\newblock In \emph{AAAI Conference on Artificial Intelligence}, 2008.

\bibitem[Marchisio et~al.(2021{\natexlab{b}})Marchisio, Koehn, and Xiong]{marchisio-etal-2021-alignment}
Kelly Marchisio, Philipp Koehn, and Conghao Xiong.
\newblock An alignment-based approach to semi-supervised bilingual lexicon induction with small parallel corpora.
\newblock In \emph{Proceedings of Machine Translation Summit XVIII: Research Track}, pages 293--304. Association for Machine Translation in the Americas, 2021{\natexlab{b}}.

\bibitem[Brown et~al.(1993)Brown, Della~Pietra, Della~Pietra, and Mercer]{brown-etal-1993-mathematics}
Peter~F. Brown, Stephen~A. Della~Pietra, Vincent~J. Della~Pietra, and Robert~L. Mercer.
\newblock The mathematics of statistical machine translation: Parameter estimation.
\newblock \emph{Computational Linguistics}, 19\penalty0 (2):\penalty0 263--311, 1993.

\bibitem[Fishkind et~al.(2019)Fishkind, Adali, Patsolic, Meng, Singh, Lyzinski, and Priebe]{Fishkind2012SeededGM}
Donniell~E. Fishkind, Sancar Adali, Heather~G. Patsolic, Lingyao Meng, Digvijay Singh, Vince Lyzinski, and Carey~E. Priebe.
\newblock Seeded graph matching.
\newblock \emph{Pattern Recognition}, 87:\penalty0 203--215, 2019.
\newblock \doi{/10.1016/j.patcog.2018.09.014}.

\bibitem[Dubossarsky et~al.(2020)Dubossarsky, Vuli{\'c}, Reichart, and Korhonen]{dubossarsky-etal-2020-secret}
Haim Dubossarsky, Ivan Vuli{\'c}, Roi Reichart, and Anna Korhonen.
\newblock The secret is in the spectra: Predicting cross-lingual task performance with spectral similarity measures.
\newblock In \emph{Proceedings of the 2020 Conference on Empirical Methods in Natural Language Processing (EMNLP)}, pages 2377--2390. Association for Computational Linguistics, 2020.
\newblock \doi{10.18653/v1/2020.emnlp-main.186}.

\bibitem[Ormazabal et~al.(2021)Ormazabal, Artetxe, Soroa, Labaka, and Agirre]{ormazabal-etal-2021-beyond}
Aitor Ormazabal, Mikel Artetxe, Aitor Soroa, Gorka Labaka, and Eneko Agirre.
\newblock Beyond offline mapping: Learning cross-lingual word embeddings through context anchoring.
\newblock In \emph{Proceedings of the 59th Annual Meeting of the Association for Computational Linguistics and the 11th International Joint Conference on Natural Language Processing (Volume 1: Long Papers)}, pages 6479--6489. Association for Computational Linguistics, 2021.
\newblock \doi{10.18653/v1/2021.acl-long.506}.

\bibitem[Zhou et~al.(2022{\natexlab{a}})Zhou, Peng, Li, and mei Han]{ZHOU2022116194}
Dong Zhou, Xiaoya Peng, Lin Li, and Jun mei Han.
\newblock Cross-lingual embeddings with auxiliary topic models.
\newblock \emph{Expert Systems with Applications}, 190, 2022{\natexlab{a}}.
\newblock \doi{10.1016/j.eswa.2021.116194}.

\bibitem[Tian et~al.(2022)Tian, Li, Ren, Zuo, Wen, Hu, Han, Huang, Deng, Zhang, and Xie]{Tian2022RAPOAA}
Zhoujin Tian, Chaozhuo Li, Shuo Ren, Zhiqiang Zuo, Zengxuan Wen, Xinyue Hu, Xiao Han, Haizhen Huang, Denvy Deng, Qi~Zhang, and Xing Xie.
\newblock {RAPO}: An adaptive ranking paradigm for bilingual lexicon induction.
\newblock In \emph{Proceedings of the 2022 Conference on Empirical Methods in Natural Language Processing}, pages 8870--8883. Association for Computational Linguistics, 2022.
\newblock \doi{10.18653/v1/2022.emnlp-main.606}.

\bibitem[JP et~al.(2020)JP, Menon, and KP]{jp-etal-2020-bucc2020}
Sanjanasri JP, Vijay~Krishna Menon, and Soman KP.
\newblock {BUCC}2020: Bilingual dictionary induction using cross-lingual embedding.
\newblock In \emph{Proceedings of the 13th Workshop on Building and Using Comparable Corpora}, pages 65--68. European Language Resources Association, 2020.

\bibitem[Huang et~al.(2020)Huang, Cai, and Church]{huang-etal-2020-improving}
Jiaji Huang, Xingyu Cai, and Kenneth Church.
\newblock Improving bilingual lexicon induction for low frequency words.
\newblock In \emph{Proceedings of the 2020 Conference on Empirical Methods in Natural Language Processing (EMNLP)}, pages 1310--1314. Association for Computational Linguistics, 2020.
\newblock \doi{10.18653/v1/2020.emnlp-main.100}.

\bibitem[Hakimi~Parizi and Cook(2021)]{hakimi-parizi-cook-2021-evaluating}
Ali Hakimi~Parizi and Paul Cook.
\newblock Evaluating a joint training approach for learning cross-lingual embeddings with sub-word information without parallel corpora on lower-resource languages.
\newblock In \emph{Proceedings of *SEM 2021: The Tenth Joint Conference on Lexical and Computational Semantics}, pages 302--307. Association for Computational Linguistics, 2021.
\newblock \doi{10.18653/v1/2021.starsem-1.29}.

\bibitem[Resiandi et~al.(2023)Resiandi, Murakami, and Nasution]{Resiandi2023NeuralNB}
Kartika Resiandi, Yohei Murakami, and Arbi~Haza Nasution.
\newblock Neural network-based bilingual lexicon induction for indonesian ethnic languages.
\newblock \emph{Applied Sciences}, 13\penalty0 (15), 2023.
\newblock \doi{10.3390/app13158666}.

\bibitem[Hakimi~Parizi and Cook(2020)]{hakimi-parizi-cook-2020-joint}
Ali Hakimi~Parizi and Paul Cook.
\newblock Joint training for learning cross-lingual embeddings with sub-word information without parallel corpora.
\newblock In \emph{Proceedings of the Ninth Joint Conference on Lexical and Computational Semantics}, pages 39--49. Association for Computational Linguistics, 2020.

\bibitem[Duong et~al.(2016)Duong, Kanayama, Ma, Bird, and Cohn]{duong-etal-2016-learning}
Long Duong, Hiroshi Kanayama, Tengfei Ma, Steven Bird, and Trevor Cohn.
\newblock Learning crosslingual word embeddings without bilingual corpora.
\newblock In \emph{Proceedings of the 2016 Conference on Empirical Methods in Natural Language Processing}, pages 1285--1295. Association for Computational Linguistics, 2016.
\newblock \doi{10.18653/v1/D16-1136}.

\bibitem[Anastasopoulos and Neubig(2020)]{anastasopoulos-neubig-2020-cross}
Antonios Anastasopoulos and Graham Neubig.
\newblock Should all cross-lingual embeddings speak {E}nglish?
\newblock In \emph{Proceedings of the 58th Annual Meeting of the Association for Computational Linguistics}, pages 8658--8679. Association for Computational Linguistics, 2020.
\newblock \doi{10.18653/v1/2020.acl-main.766}.

\bibitem[Wang et~al.(2019{\natexlab{a}})Wang, Xie, Xu, Yang, Neubig, and Carbonell]{Wang2019CrosslingualAV}
Zirui Wang, Jiateng Xie, Ruochen Xu, Yiming Yang, Graham Neubig, and Jaime~G. Carbonell.
\newblock Cross-lingual alignment vs joint training: A comparative study and a simple unified framework.
\newblock \emph{ArXiv}, abs/1910.04708, 2019{\natexlab{a}}.
\newblock \doi{10.48550/arXiv.1910.04708}.

\bibitem[Schuster et~al.(2019)Schuster, Ram, Barzilay, and Globerson]{schuster-etal-2019-cross}
Tal Schuster, Ori Ram, Regina Barzilay, and Amir Globerson.
\newblock Cross-lingual alignment of contextual word embeddings, with applications to zero-shot dependency parsing.
\newblock In \emph{Proceedings of the 2019 Conference of the North {A}merican Chapter of the Association for Computational Linguistics: Human Language Technologies, Volume 1 (Long and Short Papers)}, pages 1599--1613. Association for Computational Linguistics, 2019.
\newblock \doi{10.18653/v1/N19-1162}.

\bibitem[Zhang et~al.(2021)Zhang, Ji, Xiao, Duan, Zhang, Shi, and Luo]{zhang-etal-2021-combining}
Jinpeng Zhang, Baijun Ji, Nini Xiao, Xiangyu Duan, Min Zhang, Yangbin Shi, and Weihua Luo.
\newblock Combining static word embeddings and contextual representations for bilingual lexicon induction.
\newblock In \emph{Findings of the Association for Computational Linguistics: ACL-IJCNLP 2021}, pages 2943--2955. Association for Computational Linguistics, 2021.
\newblock \doi{10.18653/v1/2021.findings-acl.260}.

\bibitem[Li et~al.(2022)Li, Liu, Collier, Korhonen, and Vuli{\'c}]{li-etal-2022-improving}
Yaoyiran Li, Fangyu Liu, Nigel Collier, Anna Korhonen, and Ivan Vuli{\'c}.
\newblock Improving word translation via two-stage contrastive learning.
\newblock In \emph{Proceedings of the 60th Annual Meeting of the Association for Computational Linguistics (Volume 1: Long Papers)}, pages 4353--4374. Association for Computational Linguistics, 2022.
\newblock \doi{10.18653/v1/2022.acl-long.299}.

\bibitem[Li et~al.(2023)Li, Korhonen, and Vuli{\'c}]{li-etal-2023-bilingual}
Yaoyiran Li, Anna Korhonen, and Ivan Vuli{\'c}.
\newblock On bilingual lexicon induction with large language models.
\newblock In \emph{Proceedings of the 2023 Conference on Empirical Methods in Natural Language Processing}, pages 9577--9599. Association for Computational Linguistics, 2023.
\newblock \doi{10.18653/v1/2023.emnlp-main.595}.

\bibitem[Shi et~al.(2021)Shi, Zettlemoyer, and Wang]{shi-etal-2021-bilingual}
Haoyue Shi, Luke Zettlemoyer, and Sida~I. Wang.
\newblock Bilingual lexicon induction via unsupervised bitext construction and word alignment.
\newblock In \emph{Proceedings of the 59th Annual Meeting of the Association for Computational Linguistics and the 11th International Joint Conference on Natural Language Processing (Volume 1: Long Papers)}, pages 813--826. Association for Computational Linguistics, 2021.
\newblock \doi{10.18653/v1/2021.acl-long.67}.

\bibitem[Zhao and Eger(2022)]{Zhao2022ConstrainedDM}
Wei Zhao and Steffen Eger.
\newblock Constrained density matching and modeling for cross-lingual alignment of contextualized representations.
\newblock In \emph{Proceedings of Machine Learning Research 189}. ACML, 2022.

\bibitem[Lample and Conneau(2019)]{Lample2019CrosslingualLM}
Guillaume Lample and Alexis Conneau.
\newblock Cross-lingual language model pretraining.
\newblock \emph{ArXiv}, abs/1901.07291, 2019.
\newblock \doi{/10.48550/arXiv.1901.07291}.

\bibitem[Mulcaire et~al.(2019)Mulcaire, Kasai, and Smith]{mulcaire-etal-2019-polyglot}
Phoebe Mulcaire, Jungo Kasai, and Noah~A. Smith.
\newblock Polyglot contextual representations improve crosslingual transfer.
\newblock In \emph{Proceedings of the 2019 Conference of the North {A}merican Chapter of the Association for Computational Linguistics: Human Language Technologies, Volume 1 (Long and Short Papers)}, pages 3912--3918. Association for Computational Linguistics, 2019.
\newblock \doi{10.18653/v1/N19-1392}.

\bibitem[Tran et~al.(2020)Tran, Tang, Li, and Gu]{Tran2020CrosslingualRF}
C.~Tran, Y.~Tang, Xian Li, and Jiatao Gu.
\newblock Cross-lingual retrieval for iterative self-supervised training.
\newblock \emph{ArXiv}, abs/2006.09526, 2020.
\newblock \doi{10.48550/arXiv.2006.09526}.

\bibitem[Wang et~al.(2019{\natexlab{b}})Wang, Che, Guo, Liu, and Liu]{wang-etal-2019-cross}
Yuxuan Wang, Wanxiang Che, Jiang Guo, Yijia Liu, and Ting Liu.
\newblock Cross-lingual {BERT} transformation for zero-shot dependency parsing.
\newblock In \emph{Proceedings of the 2019 Conference on Empirical Methods in Natural Language Processing and the 9th International Joint Conference on Natural Language Processing (EMNLP-IJCNLP)}, pages 5721--5727. Association for Computational Linguistics, 2019{\natexlab{b}}.
\newblock \doi{10.18653/v1/D19-1575}.

\bibitem[Xu and Koehn(2021)]{Xu2021CrossLingualBC}
Haoran Xu and Philipp Koehn.
\newblock Cross-lingual bert contextual embedding space mapping with isotropic and isometric conditions.
\newblock \emph{ArXiv}, abs/2107.09186, 2021.
\newblock \doi{10.48550/arXiv.2107.09186}.

\bibitem[Wada et~al.(2021)Wada, Iwata, Matsumoto, Baldwin, and Lau]{wada-etal-2021-learning}
Takashi Wada, Tomoharu Iwata, Yuji Matsumoto, Timothy Baldwin, and Jey~Han Lau.
\newblock Learning contextualised cross-lingual word embeddings and alignments for extremely low-resource languages using parallel corpora.
\newblock In \emph{Proceedings of the 1st Workshop on Multilingual Representation Learning}, pages 16--31. Association for Computational Linguistics, 2021.
\newblock \doi{10.18653/v1/2021.mrl-1.2}.

\bibitem[Zhou et~al.(2022{\natexlab{b}})Zhou, Liu, Vuli{\'c}, Collier, and Chen]{zhou-etal-2022-prix}
Wenxuan Zhou, Fangyu Liu, Ivan Vuli{\'c}, Nigel Collier, and Muhao Chen.
\newblock Prix-{LM}: Pretraining for multilingual knowledge base construction.
\newblock In \emph{Proceedings of the 60th Annual Meeting of the Association for Computational Linguistics (Volume 1: Long Papers)}, pages 5412--5424. Association for Computational Linguistics, 2022{\natexlab{b}}.
\newblock \doi{10.18653/v1/2022.acl-long.371}.

\bibitem[Magui{\~n}o-Valencia et~al.(2018)Magui{\~n}o-Valencia, Oncevay-Marcos, and Sobrevilla~Cabezudo]{maguino-valencia-etal-2018-wordnet}
Diego Magui{\~n}o-Valencia, Arturo Oncevay-Marcos, and Marco~A. Sobrevilla~Cabezudo.
\newblock {W}ord{N}et-shp: Towards the building of a lexical database for a {P}eruvian minority language.
\newblock In \emph{Proceedings of the Eleventh International Conference on Language Resources and Evaluation ({LREC} 2018)}. European Language Resources Association (ELRA), 2018.

\bibitem[Michaud(2018)]{Michaud2018NaD}
Alexis Michaud.
\newblock Na (mosuo)-english-chinese dictionary.
\newblock In \emph{Lexica}, 2018.

\bibitem[Conneau et~al.(2020)Conneau, Khandelwal, Goyal, Chaudhary, Wenzek, Guzm{\'a}n, Grave, Ott, Zettlemoyer, and Stoyanov]{conneau-etal-2020-unsupervised}
Alexis Conneau, Kartikay Khandelwal, Naman Goyal, Vishrav Chaudhary, Guillaume Wenzek, Francisco Guzm{\'a}n, Edouard Grave, Myle Ott, Luke Zettlemoyer, and Veselin Stoyanov.
\newblock Unsupervised cross-lingual representation learning at scale.
\newblock In \emph{Proceedings of the 58th Annual Meeting of the Association for Computational Linguistics}, pages 8440--8451. Association for Computational Linguistics, 2020.
\newblock \doi{10.18653/v1/2020.acl-main.747}.

\bibitem[Lehmann et~al.(2015)Lehmann, Isele, Jakob, Jentzsch, Kontokostas, Mendes, Hellmann, Morsey, van Kleef, Auer, and Bizer]{Lehmann2015DBpediaA}
Jens Lehmann, Robert Isele, Max Jakob, Anja Jentzsch, Dimitris Kontokostas, Pablo~N. Mendes, Sebastian Hellmann, Mohamed Morsey, Patrick van Kleef, S.~Auer, and Christian Bizer.
\newblock {DB}pedia - a large-scale, multilingual knowledge base extracted from {W}ikipedia.
\newblock \emph{Semantic Web}, 6:\penalty0 167--195, 2015.

\bibitem[Vuli{\'c} et~al.(2023)Vuli{\'c}, Glava{\v{s}}, Liu, Collier, Ponti, and Korhonen]{vulicetal2022}
Ivan Vuli{\'c}, Goran Glava{\v{s}}, Fangyu Liu, Nigel Collier, Edoardo~Maria Ponti, and Anna Korhonen.
\newblock Probing cross-lingual lexical knowledge from multilingual sentence encoders.
\newblock In \emph{Proceedings of the 17th Conference of the European Chapter of the Association for Computational Linguistics}, pages 2089--2105. Association for Computational Linguistics, 2023.
\newblock \doi{10.18653/v1/2023.eacl-main.153}.

\bibitem[Reimers and Gurevych(2019)]{reimers-gurevych-2019-sentence}
Nils Reimers and Iryna Gurevych.
\newblock Sentence-{BERT}: Sentence embeddings using {S}iamese {BERT}-networks.
\newblock In \emph{Proceedings of the 2019 Conference on Empirical Methods in Natural Language Processing and the 9th International Joint Conference on Natural Language Processing (EMNLP-IJCNLP)}, pages 3982--3992. Association for Computational Linguistics, 2019.
\newblock \doi{10.18653/v1/D19-1410}.

\bibitem[El~Mekki et~al.(2023)El~Mekki, Abdul-Mageed, Nagoudi, Berrada, and Khoumsi]{el-mekki-etal-2023-promap}
Abdellah El~Mekki, Muhammad Abdul-Mageed, ElMoatez~Billah Nagoudi, Ismail Berrada, and Ahmed Khoumsi.
\newblock {P}ro{M}ap: Effective bilingual lexicon induction via language model prompting.
\newblock In \emph{Proceedings of the 13th International Joint Conference on Natural Language Processing and the 3rd Conference of the Asia-Pacific Chapter of the Association for Computational Linguistics (Volume 1: Long Papers)}, pages 577--597. Association for Computational Linguistics, 2023.
\newblock \doi{10.18653/v1/2023.ijcnlp-main.39}.

\bibitem[Erdmann et~al.(2018)Erdmann, Zalmout, and Habash]{erdmann-etal-2018-addressing}
Alexander Erdmann, Nasser Zalmout, and Nizar Habash.
\newblock Addressing noise in multidialectal word embeddings.
\newblock In \emph{Proceedings of the 56th Annual Meeting of the Association for Computational Linguistics (Volume 2: Short Papers)}, pages 558--565. Association for Computational Linguistics, 2018.
\newblock \doi{10.18653/v1/P18-2089}.

\bibitem[Bouamor et~al.(2018)Bouamor, Habash, Salameh, Zaghouani, Rambow, Abdulrahim, Obeid, Khalifa, Eryani, Erdmann, and Oflazer]{bouamor-etal-2018-madar}
Houda Bouamor, Nizar Habash, Mohammad Salameh, Wajdi Zaghouani, Owen Rambow, Dana Abdulrahim, Ossama Obeid, Salam Khalifa, Fadhl Eryani, Alexander Erdmann, and Kemal Oflazer.
\newblock The {MADAR} {A}rabic dialect corpus and lexicon.
\newblock In \emph{Proceedings of the Eleventh International Conference on Language Resources and Evaluation ({LREC} 2018)}. European Language Resources Association (ELRA), 2018.

\bibitem[Artemova and Plank(2023)]{artemova-plank-2023-low}
Ekaterina Artemova and Barbara Plank.
\newblock Low-resource bilingual dialect lexicon induction with large language models.
\newblock In \emph{Proceedings of the 24th Nordic Conference on Computational Linguistics (NoDaLiDa)}, pages 371--385. University of Tartu Library, 2023.

\bibitem[Bafna et~al.(2023)Bafna, España-Bonet, Genabith, Sagot, and Bawden]{Bafna2023ASM}
Niyati Bafna, Cristina España-Bonet, Josef Genabith, Benoît Sagot, and Rachel Bawden.
\newblock A simple method for unsupervised bilingual lexicon induction for data-imbalanced, closely related language pairs.
\newblock \emph{ArXiv}, 2023.
\newblock \doi{10.48550/arXiv.2305.14012}.

\bibitem[Ojha et~al.(2020)Ojha, Malykh, Karakanta, and Liu]{ojha-etal-2020-findings}
Atul~Kr. Ojha, Valentin Malykh, Alina Karakanta, and Chao-Hong Liu.
\newblock Findings of the {L}o{R}es{MT} 2020 shared task on zero-shot for low-resource languages.
\newblock In \emph{Proceedings of the 3rd Workshop on Technologies for MT of Low Resource Languages}, pages 33--37. Association for Computational Linguistics, 2020.
\newblock URL \url{https://aclanthology.org/2020.loresmt-1.4}.

\bibitem[Yang et~al.(2019)Yang, Luo, Chen, Liu, and Sun]{yang-etal-2019-maam}
Pengcheng Yang, Fuli Luo, Peng Chen, Tianyu Liu, and Xu~Sun.
\newblock {MAAM}: A morphology-aware alignment model for unsupervised bilingual lexicon induction.
\newblock In \emph{Proceedings of the 57th Annual Meeting of the Association for Computational Linguistics}, pages 3190--3196. Association for Computational Linguistics, 2019.
\newblock \doi{10.18653/v1/P19-1308}.

\end{thebibliography}

\end{document}